\documentclass[10pt,twocolumn,letterpaper]{article}

\usepackage[pagenumbers]{cvpr}

\usepackage[utf8]{inputenc} 
\usepackage[T1]{fontenc}    
\usepackage{graphicx}
\usepackage{amsmath}
\usepackage{amssymb}
\usepackage{booktabs}
\usepackage{enumitem}
\usepackage{colortbl} 
\usepackage{acronym}
\usepackage{balance}
\usepackage{xspace}
\usepackage{setspace}
\usepackage{tabularx,colortbl,multirow,array,makecell}
\usepackage{overpic,wrapfig}
\usepackage{nicefrac}
\usepackage[misc]{ifsym}
\usepackage{siunitx}
\usepackage{pifont}
\usepackage{dsfont}
\definecolor{cvprblue}{rgb}{0.21,0.49,0.74}
\usepackage[pagebackref,breaklinks,colorlinks,allcolors=cvprblue]{hyperref}
\usepackage[capitalize]{cleveref}
\usepackage{bm}

\definecolor{gblue}{HTML}{4285F4}
\definecolor{gred}{HTML}{DB4437}
\definecolor{ggreen}{HTML}{0F9D58}
\definecolor{vblue}{HTML}{2993ba}

\definecolor{gbest}{HTML}{FFCCCB}
\definecolor{gsecond}{HTML}{FFE5CC}
\definecolor{gthird}{HTML}{FFF2A0}

\acrodef{sdf}[SDF]{signed distance function}
\acrodef{mlp}[MLP]{multi-layer perceptron}
\acrodef{nerf}[NeRF]{Neural Radiance Field}
\acrodef{3dgs}[3D-GS]{3D Gaussain Splatting}
\acrodef{gt}[GT]{Ground Truth}
\acrodef{cd}[CD]{Chamfer Distance}
\acrodef{fscore}[F-score]{F-score}
\acrodef{nc}[NC]{Normal Consistency}
\acrodef{miou}[mIoU]{Mean Intersection over Union}
\acrodef{fr}[FR]{full-reference}
\acrodef{nr}[NR]{no-reference}
\acrodef{sds}[SDS]{Score Distillation Sampling}
\acrodef{sfm}[SfM]{Structure-from-Motion}
\acrodef{vfx}[VFX]{Visual effects}

\newcommand{\sota}{\text{state-of-the-art}\xspace}

\newcommand{\modelname}{\textbf{\textsc{DP-Recon}}\xspace}
\newcommand{\objectsdfpp}{ObjectSDF++\xspace}
\newcommand{\rico}{RICO\xspace}
\newcommand{\supp}{\textit{supplementary}\xspace}

\acrodef{cd}[CD]{Chamfer Distance}
\acrodef{nc}[NC]{Normal Consistency}
\newcommand{\fscore}{\mbox{F-Score}\xspace}

\makeatletter
\renewcommand{\paragraph}{%
  \@startsection{paragraph}{4}{\z@}%
  {1ex plus 0.5ex minus 0.2ex} 
  {-1em}                      
  {\normalfont\normalsize\bfseries} 
}
\makeatother

\def\paperID{13399}
\def\confName{CVPR}
\def\confYear{2025}

\title{Decompositional Neural Scene Reconstruction with Generative Diffusion Prior}

\author{
  Junfeng Ni\textsuperscript{1,2},
  Yu Liu\textsuperscript{1,2}, 
  Ruijie Lu\textsuperscript{2,3},
  Zirui Zhou\textsuperscript{1}, \\
  Song-Chun Zhu\textsuperscript{1,2,3},
  Yixin Chen\textsuperscript{2$\dagger$},
  Siyuan Huang\textsuperscript{2$\dagger$}
  \\
  \textsuperscript{$\dagger$} Corresponding author
  \textsuperscript{1} Tsinghua University \\
  \textsuperscript{2} State Key Laboratory of General Artificial Intelligence, BIGAI
  \textsuperscript{3} Peking University
}

\begin{document}

\twocolumn[{
\renewcommand\twocolumn[1][]{#1}
\maketitle
\vspace{-0.48in}
\begin{center}
    \centering
    \captionsetup{type=figure}
    \includegraphics[width=\linewidth]{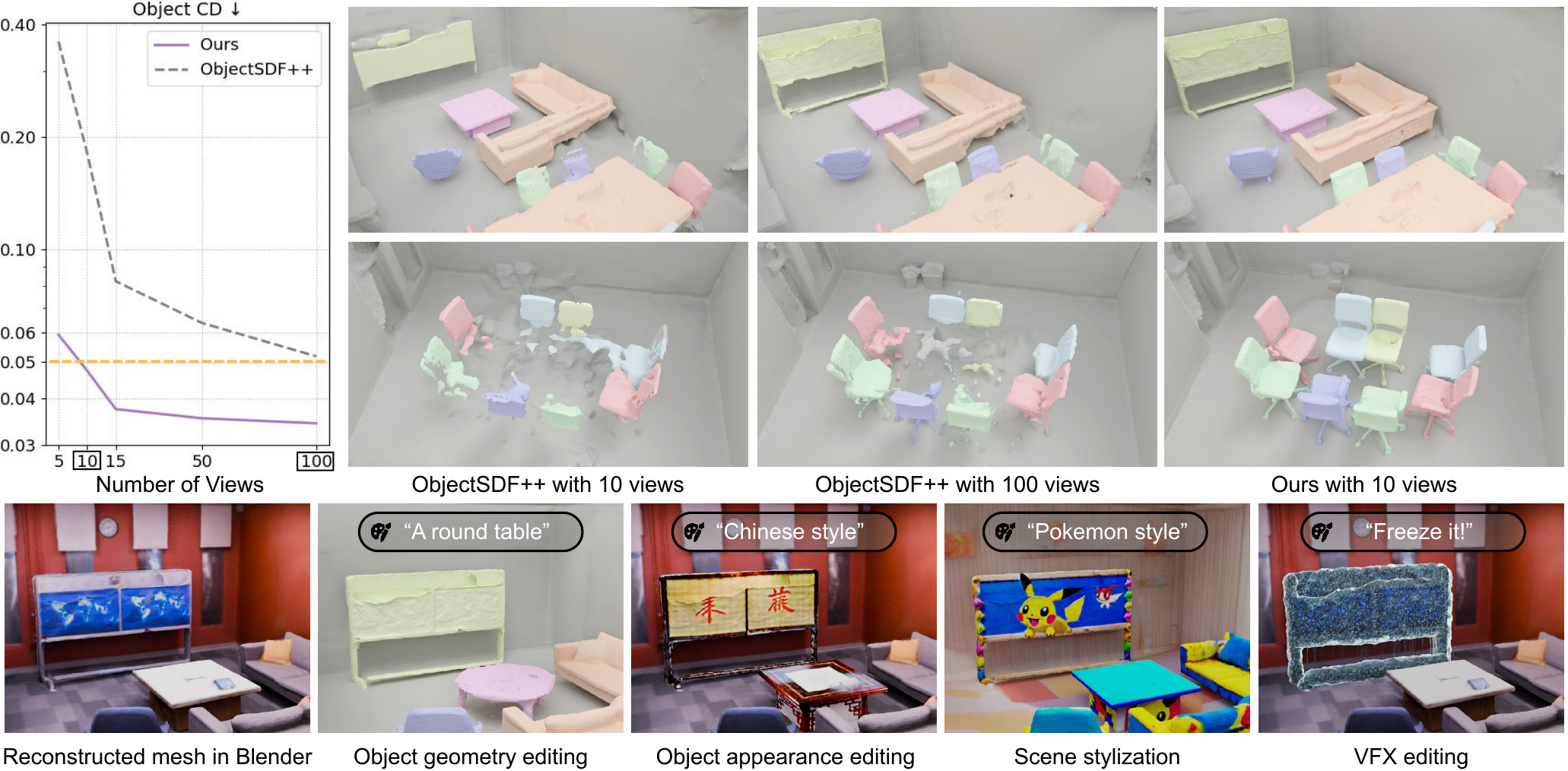}
    \vspace{-0.28in}
    \captionof{figure}{
        We propose \modelname, which capitalizes on pre-trained diffusion models for complete and decompositional neural scene reconstruction. 
        This approach significantly improves reconstruction quality in less captured regions, where previous methods often struggle. Additionally, our method enables flexible text-based editing of geometry and appearance, as well as photorealistic \acs{vfx} editing.
    }
    \label{fig:teaser}
\end{center}
}]

\begin{abstract}
Decompositional reconstruction of 3D scenes, with complete shapes and detailed texture of all objects within, is intriguing for downstream applications but remains challenging, particularly with sparse views as input. Recent approaches incorporate semantic or geometric regularization to address this issue, but they suffer significant degradation in underconstrained areas and fail to recover occluded regions. We argue that the key to solving this problem lies in supplementing missing information for these areas. To this end, we propose \modelname, which employs diffusion priors in the form of \ac{sds} to optimize the neural representation of each individual object under novel views. 
This provides additional information for the underconstrained areas, but directly incorporating diffusion prior raises potential conflicts between the reconstruction and generative guidance. Therefore, we further introduce a visibility-guided approach to dynamically adjust the per-pixel \ac{sds} loss weights. Together these components enhance both geometry and appearance recovery while remaining faithful to input images.  Extensive experiments across Replica and ScanNet++ demonstrate that our method significantly outperforms \sota methods. Notably, it achieves better object reconstruction under 10 views than the baselines under 100 views. Our method enables seamless text-based editing for geometry and appearance through \ac{sds} optimization and produces decomposed object meshes with detailed UV maps that support photorealistic \ac{vfx} editing. The project page is available at \url{https://dp-recon.github.io/}.
\end{abstract}

\section{Introduction}
\label{sec:intro}
3D scene reconstruction from multi-view images is a long-standing topic in computer vision. Recent advances in neural implicit representations~\cite{mildenhall2020nerf,kerbl3Dgaussians} have enabled significant progress in novel-view rendering~\cite{barron2022mip,yang2023freenerf,wu2023reconfusion,gao2024cat3d} and 3D geometry reconstruction~\cite{oechsle2021unisurf,wang2021neus,yariv2021volume}. Despite these advances, existing approaches are limited by representing an entire scene as a whole.
On the other hand, decompitional reconstruction~\cite{wu2022object,kong2023vmap} aims to break down the implicit 3D representation into individual objects in the scene and facilitate broader applications in embodied AI~\cite{gu2023maniskill2,anderson2018vision,krantz2020beyond,huang2023embodied,huang2025unveiling}, robotics~\cite{gong2023arnold,jiang2022vima,shridhar2022cliport,lu2024manigaussian,zhao2025tac}, and more~\cite{cui2023probio,chen2023gaussianeditor}. 

Existing methods~\cite{wu2023objsdfplus,li2023rico,ni2024phyrecon,Liu2024slotlifter} in decompositional neural reconstruction still fall short of expectations in downstream applications to reconstruct complete 3D geometry and accurate appearance (see \cref{fig:teaser}), especially in less densely captured or heavily occluded areas with sparse inputs. To address the challenge of sparse-view reconstruction, many approaches propose to incorporate semantic or geometric regularizations~\cite{jain2021putting,niemeyer2021regnerf,yang2023freenerf,kim2022infonerf}. Still, they often demonstrate significant degradation in non-observable regions since they fail to provide additional information for the underconstrained areas. Thus, we believe the key is to introduce supplementary information for these areas based on the observation from known views.

In this paper, we propose \modelname to facilitate the decompositional neural reconstruction with generative diffusion prior. Given multiple posed images, the neural implicit representation is optimized to represent both individual objects and the background within the scene. 
Besides the reconstruction loss, \textit{we employ a 2D diffusion model as a critic to supervise the optimization of each object through \ac{sds}~\cite{poole2022dreamfusion},} which iteratively refines the 3D representation by evaluating the quality of novel views from differentiable rendering.
We use the pretrained Stable Diffusion~\cite{rombach2022high}, a more general diffusion model without fine-tuning on specific datasets. We meticulously design the optimization pipeline so that the generative prior optimizes both the geometry and appearance of each object alongside the reconstruction loss, filling in the missing information in unobserved and occluded regions.

However, directly integrating the diffusion prior into the reconstruction pipeline may compromise the overall consistency, particularly in observed regions, due to their potential conflicts.  Ideally, we want to preserve the visible area in the input images while the diffusion prior completes the rest. 
To alleviate this problem, we propose a novel visibility approach that models the visibility of 3D points across the input views using a learnable grid. 
The visibility information is derived from the accumulated transmittance in volume rendering, enabling us to optimize the visibility grid without introducing computationally intensive external visibility priors~\cite{somraj2023vipnerf}. For each novel view, the visibility map can be rendered from this grid, which can dynamically adjust the per-pixel \ac{sds} and rendering loss weights, benefiting both geometry and appearance optimization stages.

Extensive experiment results on Replica~\cite{replica19arxiv} and ScanNet++~\cite{yeshwanthliu2023scannetpp} demonstrate that our method significantly surpasses all \sota methods in both geometry and appearance reconstruction, particularly in heavily occluded regions. \textit{Remarkably, with only 10 input views, our method achieves object reconstruction quality superior to baseline methods that rely on 100 input views for heavily occluded scenes in ~\cref{fig:teaser}.} The ablative studies highlight the effectiveness of incorporating generative diffusion prior with visibility guidance. Our method enables seamless scene-level and object-level editing, \eg, geometry and appearance stylization, using \ac{sds} optimization. It produces decomposed object meshes with detailed UV maps, enabling photorealistic rendering and \ac{vfx} editing in common 3D software, thereby supporting various downstream applications.

In summary, our main contributions are three-fold:
\begin{itemize}[leftmargin=*]
    \item We introduce a novel method \modelname that incorporates generative prior into decompositional scene reconstruction, significantly improving geometry and appearance recovery, particularly in heavily occluded regions.
    \item We propose a visibility-guided approach to dynamically adjust the \ac{sds} loss, alleviating the conflict between the reconstruction objective and generative prior guidance.
    \item Extensive experiments demonstrate that our model significantly enhances both geometry and appearance. Our method enables seamless geometry and appearance editing, yielding decomposed object meshes with detailed UV maps for broad downstream applications.
\end{itemize}

\section{Related Work}
\label{sec:related_work}
\subsection{Neural Implicit Surface Reconstruction}
Recent advances in neural implicit representations~\cite{nie2020cvpr,Zhang_2021_im3d,liu2022towards,chen2023ssr,niemeyer2020differentiable,chibane2020implicit,martel2021acorn} has inspired the efforts to bridge the volume density in \ac{nerf}~\cite{mildenhall2020nerf,zhang2020nerf++} with iso-surface representations, \eg, occupancy~\cite{mescheder2019occupancy} or \ac{sdf}~\cite{park2019deepsdf,oechsle2021unisurf,wang2021neus,yariv2021volume}, enabling reconstruction from 2D images. To facilitate practical applications like scene editing~\cite{wang2024roomtex,tang2024intex} and manipulation~\cite{li2024ag2manip,li2024grasp}, advanced methods~\cite{wu2022object,kong2023vmap,lyu2024total} target compositional scene reconstruction by decomposing the implicit 3D representation into the individual objects, incorporating additional information in semantics~\cite{wu2023objsdfplus,li2023rico} or physics simulation~\cite{ni2024phyrecon,yang2024physcene}. While these methods realize plausible object disentanglement, they still face significant challenges in recovering complete objects and smooth backgrounds, especially in regions unobserved from the input images, such as uncaptured areas or objects behind occlusion. In this paper, we aim to enhance the reconstruction quality of both geometry and appearance, recovering the objects in complete shape and texture for more versatile downstream applications~\cite{jiang2024scaling,jiang2024autonomous,liu2025building,gong2023arnold}.

\subsection{Sparse-view \ac{nerf}}
\acp{nerf}~\cite{mildenhall2020nerf,barron2022mip,barron2023zip}, while demonstrating impressive results, rely on hundreds of posed images during training to effectively optimize the 3D representations. Many works have attempted to reduce \ac{nerf}'s reliance on dense image capture through various regularization techniques~\cite{somraj2023vipnerf,sparf2023,seo2023cvpr,somraj2023simplenerf,nerfbusters2023,wei2021nerfingmvs,kwak2023geconerf,hu2023consistentnerf,younes2024sparsecraft,ye2024pvprecon,sun2024p2nerf}. For example, RegNeRF~\cite{niemeyer2021regnerf} uses a depth smoothness loss alongside normalizing flow to regularize both geometry and appearance in novel views. Other methods incorporate depth information from \ac{sfm}~\cite{kangle2021dsnerf,roessle2022depthpriorsnerf} or monocular estimators~\cite{yu2022monosdf,wang2022sparsenerf}. DietNeRF~\cite{jain2021putting} improves novel view geometry through cross-view semantic consistency, while FreeNeRF~\cite{yang2023freenerf} reduces artifacts in novel views by regularizing the frequency of \ac{nerf}'s positional encoding features.
While these methods yield plausible results in regions \textit{with limited image coverage}, they still fail in areas\textit{ with no captured observations}. We argue the key to addressing this issue lies in introducing external knowledge, \ie, from pretrained diffusion models, and harmonizing generative and reconstruction guidance.

\begin{figure*}[htbp]
    \centering
    \includegraphics[width=\linewidth]{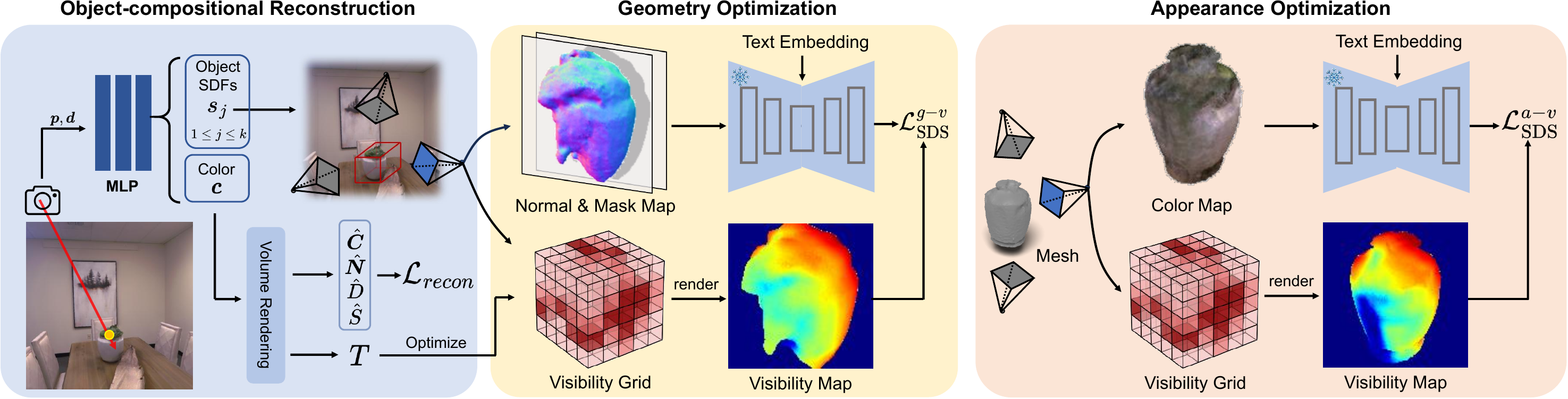}
    \caption{\textbf{Overview of \modelname.} 
   We first use reconstruction loss \(\mathcal{L}_{recon}\) for decompositional neural reconstruction, followed by the prior-guided geometry optimization stage that incorporates \acs{sds} loss $\mathcal{L}_{\text{SDS}}^{g-v}$. We finally export the object meshes and optimize their appearance with $\mathcal{L}_{\text{SDS}}^{a-v}$. The visibility
   balances the guidance from prior and reconstruction by dynamically adjusting per-pixel \acs{sds} loss.}
    \label{fig:method}
    \vspace{-2mm}
\end{figure*}

\subsection{Diffusion Prior for 3D Reconstruction}
Recently, diffusion models have proven effective in providing prior knowledge for reconstruction~\cite{zhou2023sparsefusion,liu2023zero1to3,shi2023MVDream,chen2023fantasia3d,qiu2024richdreamer,lu2024movis}. DreamFusion~\cite{poole2022dreamfusion} introduces \ac{sds} with Stable Diffusion~\cite{rombach2022high} to guide 3D object generation from text prompts. 
Methods such as ReconFusion~\cite{wu2023reconfusion}, NeRFiller~\cite{weber2023nerfiller}, MVIP-NeRF~\cite{chen2024mvip}, and ExtraNeRF~\cite{shih2024extranerf} refine fine-tuned 2D diffusion models to recover or inpaint high-fidelity \ac{nerf} from sparse input views.
More recent approaches leverage video diffusion models~\cite{melaskyriazi2024im3d,gao2024cat3d,liu2024reconx,liu20243dgsenhancer,lu2025taco} for improved consistency across views. 
However, these methods only focus on the novel view synthesis by applying diffusion priors to the entire scene, without awareness of individual objects. While the results seem reasonable, they fail to maintain the 3D consistency of objects across views, do not recover the 3D geometry of objects and cannot reconstruct regions behind occlusion. On the contrary, our method leverages the benefits of decompositional scene reconstruction and applies a generative prior to each object. This substantially enhances the reconstruction quality of both individual objects and the overall scene, in terms of both geometry and appearance. 
Moreover, we identify a critical issue overlooked in prior work: the conflict between generative and reconstruction guidance, and introduce a novel visibility-guided strategy to dynamically adjust the \ac{sds} loss during training, effectively resolving this conflict.

\section{Method}
\label{sec:method}
Given a set of posed RGB images and corresponding instance masks, we aim to reconstruct the geometry and appearance of objects and the background in the scene. \cref{fig:method} presents an overview of our proposed \modelname .

\subsection{Background}
\label{sec:background}
\paragraph{SDF-based Neural Implicit Surfaces}
\ac{nerf}~\cite{mildenhall2020nerf} learns the density field \(\sigma(\bm{p})\) and color field \(\bm{c}(\bm{p},\bm{d})\) from input images for each point \(\bm{p}\) and view direction \(\bm{d}\). To reconstruct the 3D geometric surface, current approaches~\cite{wang2021neus,yariv2021volume} replace the \ac{nerf} density \(\sigma(\bm{p})\) by a learnable transformation function of \ac{sdf} value \(s(\bm{p})\), which is defined as the signed distance from point \(\bm{p}\) to the boundary surface. Following MonoSDF~\cite{yu2022monosdf}, for a ray \(\bm{r}\) with view direction \(\bm{d}\), we use the SDF \(s(\bm{p}_{i})\) and the color \(\bm{c}(\bm{p}_{i},\bm{d})\) for each point \(\bm{p}_{i}\) along the ray to render the color \(\hat{\bm{C}}(\bm{r})\) by volume rendering:
\begin{equation}
\label{eq:volume_rendering_color}
    \hat{\bm{C}}(\bm{r}) = \sum_{i=0}^{n-1} T_{i}\alpha_{i}\bm{c}_{i},
\end{equation}
where \(T_{i}\) is the discrete accumulated transmittance, and \(\alpha_{i}\) is the discrete opacity value, defined as:
\begin{equation}
\label{eq:transmittance}
    T_{i} = \prod_{j=0}^{i-1}(1-\alpha_{j}),
    \quad \alpha_i = 1-exp(-\sigma_i\delta_i),
\end{equation}
where \(\delta_i\) represents the distance between neighboring sample points along the ray. Depth \(\hat{D}(\bm{r})\) and normal \(\hat{\bm{N}}(\bm{r})\) can also be derived through volume rendering.

\paragraph{Object-compositional Scene Reconstruction}
Following previous work~\cite{wu2022object,wu2023objsdfplus,li2023rico,ni2024phyrecon}, we consider the decompositional reconstruction of objects utilizing their corresponding masks and treat the background as an object. Specifically, for a scene with \(k\) objects, we predict \(k\) SDFs for each point \(\bm{p}\) and the \(j\)-th \( (1 \leq j \leq k) \) SDF \(s_j(\bm{p})\) is for the \(j\)-th object. The scene \ac{sdf} \(s(\bm{p})\) is the minimum of the object SDFs. We use the SDF \(s_j(\bm{p})\) and the color \(\bm{c}(\bm{p}_{i},\bm{d})\) to render color \(\hat{\bm{C}}_j(\bm{r})\), depth \(\hat{D}_j(\bm{r})\) and normal \(\hat{\bm{N}}_j(\bm{r})\) for \(j\)-th object. See more details in the \supp.

\paragraph{Score distillation sampling}
DreamFusion~\cite{poole2022dreamfusion} enables the optimization of any differentiable image generator, \eg 3D \ac{nerf}, from textual descriptions by employing a pre-trained 2D diffusion model~\cite{saharia2022imagen,rombach2022high}. Formally, let \(x=g(\theta)\) represent an image rendered by a differentiable generator \(g\) with parameter \(\theta\), DreamFusion leverages a diffusion model \(\phi\) to provide a score function \(\hat{\epsilon}_{\phi}(x_t;y,t)\), which predicts the sampled noise \(\epsilon\) given the noisy image \(x_t\), text-embedding \(y\), and noise level \(t\). This score function guides the direction of the gradient for updating the parameter \(\theta\), and the gradient is calculated by \acf{sds}:

\begin{equation}
\nabla_\theta \mathcal{L}_{\text{SDS}}(\phi, x) = \mathbb{E}_{t, \epsilon} \left[ w(t) \left( \hat{\epsilon}_\phi(x_t; y, t) - \epsilon \right) \frac{\partial x}{\partial \theta} \right],
\end{equation}
where \(w(t)\) is a weighting function.

\subsection{3D Reconstruction with Generative Priors}
\label{sec:generative_prior}
The latent neural representation of the 3D scene is primarily optimized by the reconstruction loss \(\mathcal{L}_{recon}\) derived from volume rendering, following prior work~\cite{wu2022object,wu2023objsdfplus,li2023rico}. However, regions with sparse capture or heavy occlusions often lead to suboptimal geometry and appearance recovery due to insufficient information as reconstruction guidance. To mitigate this gap, we introduce diffusion prior to optimize the the 3D model, both in geometry and appearance, so that it looks realistic at novel unobserved views.

\paragraph{Prior-guided Geometry Optimization}
We adopt the decompositional neural implicit surface as our 3D representation, which is parameterized with a series of \acp{mlp} with parameter $\theta$. The rendering functions in \cref{sec:background} serve as the image generator \(g(\theta)\). At each training iteration, we sample the \(j\)-th object and render its normal map and mask map at a randomly sampled camera pose. Following previous work~\cite{chen2023fantasia3d,qiu2024richdreamer}, we use a concatenated map $\tilde{n}_j$ of the normal and mask maps as the input for the diffusion model to improve geometric optimization stability. We then employ the \ac{sds} loss to compute the gradient for updating \(\theta\) as follows:

\begin{equation}
\label{eq:geo_sds}
\nabla_\theta \mathcal{L}_{\text{SDS}}^{g} = \mathbb{E}_{t, \epsilon} \left[ w(t) \left( \hat{\epsilon}_\phi(z_t; y, t) - \epsilon \right) \frac{\partial z}{\partial \tilde{n}_j} \frac{\partial \tilde{n}_j}{\partial \theta} \right],
\end{equation}
where \(z\) is the latent code of \(\tilde{n}_j\). The background is also treated as one object for geometry optimization. 

\paragraph{Prior-guided Appearance Optimization}
To produce object meshes with detailed UV maps, which are friendly for photorealistic rendering in common 3D software and enable more downstream applications, we directly optimize the mesh appearance rather than \ac{nerf}'s appearance field. More specifically, we export the mesh for each object after the geometry optimization stage. Using NVDiffrast~\cite{Laine2020diffrast} for differentiable mesh rendering, we employ another small network \(\psi\) to predict color for the mesh surface points. At each training iteration, the color map \(c_j\) for \(j\)-th is rendered at a randomly selected camera view, and the appearance \ac{sds} loss is used to compute the gradient for updating \(\psi\):
\begin{equation}
\label{eq:color_sds}
\nabla_\psi \mathcal{L}_{\text{SDS}}^{a} = \mathbb{E}_{t, \epsilon} \left[ w(t) \left( \hat{\epsilon}_\phi(z_t; y, t) - \epsilon \right) \frac{\partial z}{\partial c_j} \frac{\partial c_j}{\partial \psi} \right],
\end{equation}
where \(z\) is the latent code of \(c_j\). Note that the color rendering loss from input views is also used to optimize \(\psi\).

\paragraph{Background Appearance Optimization}
Applying appearance \ac{sds} in \cref{eq:color_sds} for background optimization can lead to degenerated results, \eg, introducing non-existent objects, as the background lacks clear geometric cues from the local camera perspective. To mitigate this shape-appearance ambiguity, we use depth-guided inpainting~\cite{zhang2023adding} for the background panorama color map and employ the inpainted panorama to supervise background color during the appearance optimization stage. The inpainting mask is based on the visibility of the pixel in the panorama, derived from our visibility modeling introduced in \cref{sec:visibility}.

\subsection{Visibility-guided Optimization}
\label{sec:visibility}
\acf{sds}, despite its wide application, has been shown to suffer from significant artifacts~\cite{yu2023text,lee2024dreamflow}, such as oversaturation, oversmoothing, and low-diversity, and optimization instability~\cite{wang2024prolificdreamer,mcallister2024rethinking}. They become even more significant when optimizing the latent 3D representation through both reconstruction and \ac{sds} guidance, due to their potential conflict, leading to inconsistencies with the observations. We address this problem by proposing a visibility-guided approach, which adjusts geometry and appearance \ac{sds} loss based on pixel visibility in the input view when rendered from a novel view.

\paragraph{Visibility Modeling}
We introduce a learnable visibility grid \(G\) to model the visibility \(v\) of a 3D point \(\bm{p}\) in the input views. We employ a view-independent modeling for visibility, \ie, \(v=G(\bm{p})\), as it only depends on the input views and is independent of the ray direction from novel views.

\begin{table*}[ht!]
\small
\caption{\textbf{Quantitative results on reconstruction and novel view synthesis.} Our method achieves superior or comparable reconstruction and rendering quality compared to the baselines. This highlights the effectiveness of incorporating a generative prior to improve overall reconstruction quality, while the visibility modeling ensures stability in the observable parts, preventing drastic changes.}
\label{tab:quantitative_scene_results}
\centering
\begin{tabular}{llccccccc}
\toprule
\multirow{2}{*}{Dataset}   & \multirow{2}{*}{Method} & \multicolumn{3}{c}{Reconstruction}   & \multicolumn{4}{c}{Rendering} \\
\cmidrule(lr){3-5} \cmidrule(lr){6-9}
                                               &                         & \acs{cd}$\downarrow$                    & \fscore$\uparrow$               & \acs{nc}$\uparrow$                     & PSNR$\uparrow$  & SSIM$\uparrow$  & LPIPS$\downarrow$ & MUSIQ$\uparrow$ \\ 
\midrule
\multirow{7}{*}{Replica}    & ZeroNVS*                & 21.53 & 16.41 & 79.43 & 14.47 & 0.515 & 0.428 & \cellcolor{gsecond}45.78 \\
                            & FreeNeRF                & 67.75 & 6.63  & 48.59 & 13.69 & 0.437 & 0.513 & 37.54 \\
                            & MonoSDF                 & \cellcolor{gthird}12.57 & \cellcolor{gthird}43.25  & \cellcolor{gthird}83.14                  & 22.44 & 0.809 & \cellcolor{gthird}0.246 & 36.02 \\
                            & RICO                    & 17.36                 & 27.89                 & 82.27                  & 19.85 & 0.746 & 0.356 & 31.82 \\
                            & ObjectSDF++             & \cellcolor{gsecond}8.57                 & \cellcolor{gsecond}50.11                 & \cellcolor{gsecond}85.44                  & \cellcolor{gsecond}24.66 & \cellcolor{gsecond}0.865 & \cellcolor{gsecond}0.198 & 41.42 \\
                            & Ours$_{geo}$            & \cellcolor{gbest}7.91 & \cellcolor{gbest}50.99 & \cellcolor{gbest}89.36 & \cellcolor{gbest}25.08 & \cellcolor{gbest}0.868 & \cellcolor{gbest}0.196 & \cellcolor{gthird}43.33 \\
                            & Ours                    & \cellcolor{gbest}7.91 & \cellcolor{gbest}50.99 & \cellcolor{gbest}89.36 & \cellcolor{gthird}24.52 & \cellcolor{gthird}0.846 & 0.286 & \cellcolor{gbest}49.22 \\
\midrule
\multirow{7}{*}{ScanNet++} & ZeroNVS*                & 36.69 & 6.48 & 69.61 & 12.22      & 0.463      & 0.443      & \cellcolor{gsecond}41.36      \\
\multicolumn{1}{l}{}       & FreeNeRF                & 134.34 & 1.50 & 46.46 & 15.32      & 0.533      & 0.481      & 35.42      \\
\multicolumn{1}{l}{}       & MonoSDF                 & \cellcolor{gthird}18.52 & \cellcolor{gthird}26.72 & \cellcolor{gthird}76.26 & 17.58 & 0.646 & 0.451 & 28.63      \\
\multicolumn{1}{l}{}       & RICO                    & 23.64 & 21.28 & 65.28 & 15.74 & 0.616 & 0.467 & 26.95      \\
\multicolumn{1}{l}{}       & ObjectSDF++             & \cellcolor{gsecond}10.67 & \cellcolor{gsecond}44.78 & \cellcolor{gsecond}78.18 & \cellcolor{gthird}19.43 & \cellcolor{gsecond}0.741 & \cellcolor{gsecond}0.332 & 33.64      \\
\multicolumn{1}{l}{}       & Ours$_{geo}$                    & \cellcolor{gbest}10.18  & \cellcolor{gbest}45.13 & \cellcolor{gbest}81.87 & \cellcolor{gsecond}20.13 & \cellcolor{gbest}0.752 & \cellcolor{gbest}0.319 & \cellcolor{gthird}38.43      \\ 
\multicolumn{1}{l}{}       & Ours           & \cellcolor{gbest}10.18  & \cellcolor{gbest}45.13 & \cellcolor{gbest}81.87 & \cellcolor{gbest}20.17 & \cellcolor{gthird}0.715 & \cellcolor{gthird}0.442 & \cellcolor{gbest}45.71      \\
\bottomrule
\end{tabular}
\end{table*}

\begin{figure*}[ht!]
    \centering
    \includegraphics[width=0.9\linewidth]{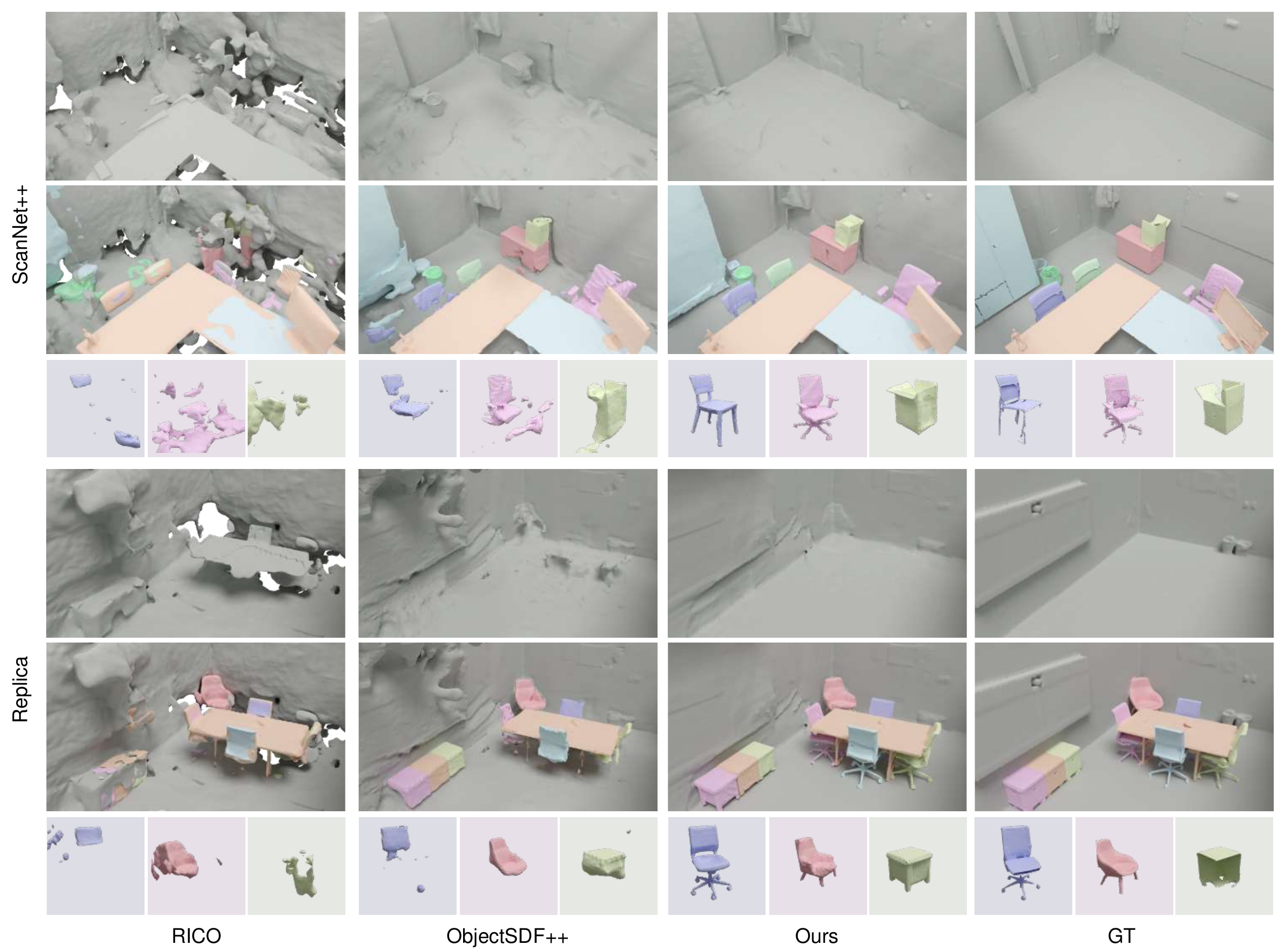}
    \caption{\textbf{Qualitative comparison of 10-views reconstruction.} We present examples from ScanNet++~\cite{yeshwanthliu2023scannetpp} and Replica~\cite{replica19arxiv}. In each example, the first row shows the background, the second the full scene, and the third individual objects. We reconstruct more complete and reasonable 3D geometry, especially in less captured and occluded regions, such as the chair behind the table and the background.}
    \label{fig:mesh_results}
\end{figure*}

\begin{figure*}[ht!]
    \centering
    \includegraphics[width=\linewidth]{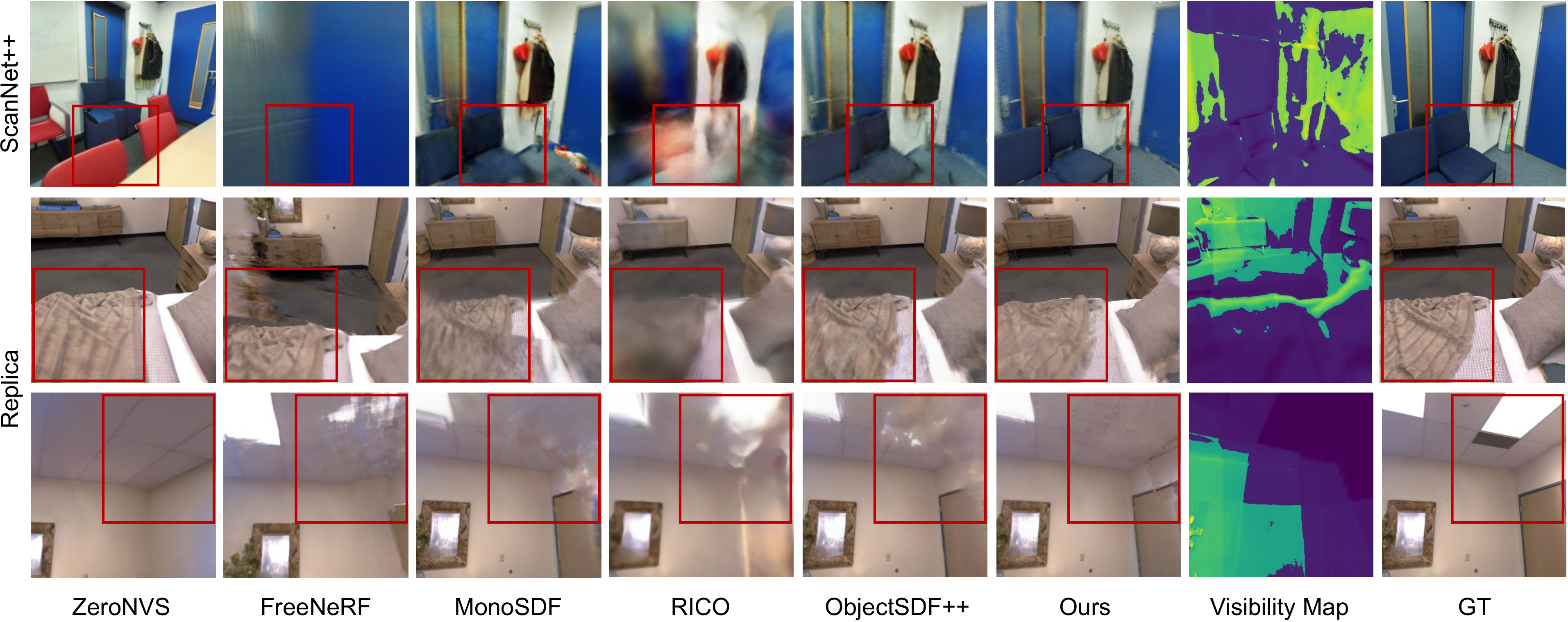}
    \caption{\textbf{Qualitative results of novel view synthesis}. Our method significantly improves rendering quality, particularly in less captured regions with low visibility, shown in darker colors in the visibility maps, such as the highlighted corner of the wall. }
    \label{fig:color_results}
    \vspace{-2mm}
\end{figure*}
Ideally, points observed in more input views should have higher visibility. The accumulated transmittance \(T\) for a 3D point \(\bm{p}\) represents the probability that the corresponding ray reaches \(\bm{p}\) without hitting any other particles - higher transmittance \(T\) means greater visibility probability in the input views. Therefore, we initialize \(G\) as zero and utilize the \(T\) from input views to optimize the visibility grid \(G\) via:
\begin{equation}
\label{eq:visibility_loss}
    \mathcal{L}_{v} = \sum_{i=0}^{n} max(T_i - G(p_i), 0).
\end{equation}
We detach the gradient of \(T_i\) to avoid the influence on the reconstruction network.
We optimize \(G\) after finishing the decompositional reconstruction stage to ensure the accuracy of the transmittance and freeze \(G\) in the geometry and appearance optimization stage with generative diffusion prior.

\paragraph{Visibility-guided \ac{sds}}
We obtain the visibility map \(V\) under novel view by volume rendering similar to \cref{eq:volume_rendering_color}. \(V\) for a ray \(r\) is calculated as $V(r) = \sum_{i=0}^{n-1} T_{i}\alpha_{i}v_{i}$.
The visibility weighting function \(w^{v}(z)\) is calculated as:
\begin{equation}
    w^{v}(z) = 
    \begin{cases} 
     w_0 + m_0 V(z) & \text{if } V(z) \leq \tau \\ 
     w_1 + m_1 V(z) & \text{if } V(z) > \tau
    \end{cases},
\end{equation}
where \(w\) and \(m\) are piecewise linear coefficients, \(V(z)\) denotes the pixel-wise visibility associated with latent \(z\), and \(\tau\) a threshold separating high and low visibility area.
We reduce the \ac{sds} loss weight in high visibility regions to enhance reconstruction guidance while increasing \ac{sds} loss weight in low visibility regions for higher generative prior guidance.
Then we rewrite \cref{eq:geo_sds} and \cref{eq:color_sds} as:
\begin{equation}
\label{eq:visibility_guided_sds}
\begin{split}
    \scalebox{0.95}{$\nabla_\theta \mathcal{L}_{\text{SDS}}^{g-v} = \mathbb{E}_{t, \epsilon} \left[ w^{v}(z) w(t) \left( \hat{\epsilon}_\phi(z_t; y, t) - \epsilon \right)  \frac{\partial z}{\partial n_j} \frac{\partial n_j}{\partial \theta}  \right] $} \\
    \scalebox{0.95}{$\nabla_\psi \mathcal{L}_{\text{SDS}}^{a-v} = \mathbb{E}_{t, \epsilon} \left[ w^{v}(z) w(t) \left( \hat{\epsilon}_\phi(z_t; y, t) - \epsilon \right) \frac{\partial z}{\partial c_j} \frac{\partial c_j}{\partial \psi} \right] $}
\end{split}
\end{equation}

\subsection{To Make Prior-Guided Optimization Work}
\paragraph{Training process}
Our training consists of three stages: 

\paragraph{(1) Object-compositional reconstruction} The implicit surfaces are optimized to decompose the scene into individual objects with the reconstruction loss \(\mathcal{L}_{recon}\) following prior work~\cite{wu2023objsdfplus,li2023rico}. See more details in \supp. After this stage, we optimize the visibility grid \(G\) by \(\mathcal{L}_{v}\) in \cref{eq:visibility_loss} and keep it frozen in the following two stages.

\paragraph{(2) Geometry optimization} In addition to \(\mathcal{L}_{recon}\), we also apply visibility-guided geometry \ac{sds} \(\mathcal{L}_{\text{SDS}}^{g-v}\) for each object to optimize the latent representation.

\paragraph{(3) Appearance optimization} We export the mesh for each object after the geometry optimization and optimize the appearance network \(\psi\) with \(\mathcal{L}_{\text{SDS}}^{a-v}\) and color rendering loss. The appearance of the background is additionally reconstructed with the inpainted panorama.

\paragraph{Effective Rendering for \acs{sds}}
As introduced above, \(\mathcal{L}_{\text{SDS}}^{g-v}\) requires the normal and mask maps from volume rendering for iterative optimization. However, traditional volume rendering is slow, \eg, VolSDF~\cite{yariv2021volume} takes about 0.5 seconds to render a full image at \(128 \times 128\) resolution, which is impractical for optimization. To address this, we apply the OccGrid sampling method~\cite{li2023nerfacc} to render normal map and mask map for \ac{sds} novel views, reducing rendering time to only 0.01 seconds for a \(128 \times 128\) resolution image.

\paragraph{Novel View Selection}
\ac{sds} optimization requires novel view images rendered under object-centric camera poses. However, due to insufficient constraints with sparse views, \ac{nerf}-based methods produce floating artifacts throughout the scene, making it difficult to render object-centric images for each object. To address this, we use visibility grid \(G\) to predict the boundary of each object and filter out the floaters. After filtering, we obtain the object's bounding box and sample novel views for \ac{sds} around it. For the background, we randomly sample novel views within the scene. 

\begin{table}[t!]
\small
\centering
\caption{\textbf{Decompositional object reconstruction.} Our approach significantly outperforms baselines, recovering complete object meshes and smoother backgrounds with generative prior.}
\label{tab:quantitative_object_results}
\resizebox{\linewidth}{!}{
\begin{tabular}{llccccccc}
\toprule
\multicolumn{1}{l}{\multirow{2}{*}{Method}} & \multicolumn{4}{c}{Object Reconstruction} & \multicolumn{3}{c}{BG Reconstruction} \\ \cmidrule(lr){2-5} \cmidrule(lr){6-8} 
                           & \acs{cd}$\downarrow$      & \fscore$\uparrow$      & \acs{nc}$\uparrow$     & mIoU$\uparrow$     & \acs{cd}$\downarrow$        & \fscore$\uparrow$        & \acs{nc}$\uparrow$       \\ \midrule
\rowcolor{gray!30} \multicolumn{8}{l}{\textbf{Replica}} \\ 
                           RICO                                         & 10.32  & 49.26       & 61.27 & 71.21   & 13.35   & 39.73         & 85.32   \\
                           ObjectSDF++                                  & 7.49  & 56.69       & 64.75 & 71.72  & 10.33   & 44.19         & 86.34   \\
                           Ours                                         & \cellcolor{gbest}5.54 & \cellcolor{gbest}67.71       & \cellcolor{gbest}73.50 & \cellcolor{gbest}88.21 & \cellcolor{gbest}9.39 & \cellcolor{gbest}46.14 & \cellcolor{gbest}92.83   \\ \midrule
\rowcolor{gray!30} \multicolumn{8}{l}{\textbf{ScanNet++}} \\ 
                           RICO                                         & 24.09  & 39.26       & 58.26 & 42.25   & 18.37          & 34.72               & 78.26         \\
                           ObjectSDF++                                  & 14.52  & 46.87       & 61.57 & 45.73   & 13.20          & 38.92               & 80.47         \\
                           Ours                                         & \cellcolor{gbest}5.03 & \cellcolor{gbest}66.55 & \cellcolor{gbest}72.91 & \cellcolor{gbest}70.01 & \cellcolor{gbest}11.51 & \cellcolor{gbest}40.12 & \cellcolor{gbest}86.24 \\
                           \bottomrule
\end{tabular}}
\end{table}

\section{Experiments}
We evaluate \modelname on both geometry and appearance recovery for sparse-view 3D reconstruction.
Additionally, we provide generalization results on YouTube videos, failure cases, and discuss limitations in \supp.

\subsection{Settings}
\paragraph{Datasets}
We conduct experiments on the synthetic dataset Replica~\cite{replica19arxiv} with 8 scenes following MonoSDF~\cite{yu2022monosdf} and ObjectSDF++~\cite{wu2023objsdfplus}. Additionally, we use 6 scenes from the real-world dataset ScanNet++~\cite{yeshwanthliu2023scannetpp}. We use 10 input views for each scene, except experiments on different input view numbers in \cref{tab:view_results}. See more details in \supp.

\paragraph{Baselines}
We compare against the state-of-tart sparse-view \ac{nerf} method FreeNeRF~\cite{yang2023freenerf} and dense-view MonoSDF~\cite{yu2022monosdf} with geometric regularization. 
We also compare with \rico~\cite{li2023rico} and \objectsdfpp~\cite{wu2023objsdfplus} for decompositional reconstruction.
We adapt ZeroNVS~\cite{Kyle2024zeronvs}, which synthesizes novel view of scenes from a single image, following ReconFusion~\cite{wu2023reconfusion} to multiview settings. 

\begin{table*}[ht!]
    \small
    \centering
    \caption{\textbf{Quantitative results on different view numbers.} Our method outperforms baselines significantly under sparse-view settings.}
    \label{tab:view_results}
    \resizebox{\linewidth}{!}{
    \begin{tabular}{lccccccccc}
        \toprule
        & \multicolumn{3}{c}{Scene Reconstruction (CD$\downarrow$ / NC$\uparrow$)} & \multicolumn{3}{c}{Object Reconstruction (CD$\downarrow$ / mIoU$\uparrow$)} & \multicolumn{3}{c}{Rendering (PSNR$\uparrow$/MUSIQ$\uparrow$)} \\
        \cmidrule(lr){2-4} \cmidrule(lr){5-7} \cmidrule(lr){8-10}
        Method & 5 views & 10 views & 15 views & 5 views & 10 views & 15 views & 5 views & 10 views & 15 views \\
        \midrule
        ZeroNVS* & 31.86 / 75.26 & 21.53 / 79.43 & 20.80 / 81.81 & - & - & - & 13.72 / 39.01 & 14.47 / 45.78 & 15.02 / 46.39 \\
        FreeNeRF & 76.83 / 48.19 & 67.76 / 48.59 & 65.26 / 49.71  & - & - & - & 12.94 / 27.71 & 13.69 / 37.54 & 13.89 / 37.13 \\
        MonoSDF & 31.26 / 76.03 & 12.57 / 83.14 & 8.94 / 87.23 & - & - & - & 18.57 / 31.47 & 22.44 / 36.02 & 26.20 / 42.25 \\
        RICO & 37.83 / 72.30 & 17.36 / 82.27 & 8.84 / 86.28 & 15.81 / 45.27 & 10.32 / 71.21 & 8.43 / 78.04 & 16.78 / 28.69 & 19.85 / 31.82 & 20.57 / 30.81 \\
        ObjectSDF++ & 35.49 / 68.86 & 8.57 / 85.44 & 7.21 / 88.21 & 9.35 / 47.63 & 7.49 / 71.72 & 6.93 / 80.01 & 17.43 / 28.13 & \textbf{24.66} / 41.42 & 27.33 / 44.36 \\
        Ours & \textbf{12.63} / \textbf{83.72} & \textbf{7.91} / \textbf{89.36} & \textbf{6.78} / \textbf{91.79} & \textbf{6.66} / \textbf{69.48} & \textbf{5.54} / \textbf{88.21} & \textbf{4.88} / \textbf{87.39} & \textbf{20.43} / \textbf{39.28} & 24.52 / \textbf{49.22} & \textbf{28.12} / \textbf{50.71} \\
        \bottomrule
    \end{tabular}
    }
\end{table*}

\paragraph{Metrics}
For reconstruction metrics, we evaluate \ac{cd}, \fscore, and \ac{nc} following MonoSDF~\cite{yu2022monosdf} for both total scene and decompositional reconstruction. Additionally, we assess novel view mask \ac{miou} to evaluate the completeness of object reconstruction.
For rendering metrics, we evaluate \ac{fr} and \ac{nr} following ExtraNeRF~\cite{shih2024extranerf}. For the \ac{fr} metrics, we use PSNR, SSIM, and LPIPS and for \ac{nr}, we employ MUSIQ~\cite{ke2021musiq} to evaluate the visual quality of rendered images. 
We randomly sample 10 novel views within each scene to evaluate the rendering metrics and mask \ac{miou}.

\subsection{Results}

\begin{table}[ht!]
    \small
    \centering
    \caption{\textbf{Ablation study.} Geometry (\textit{GP}) and appearance (\textit{AP}) priors improve the reconstruction and rendering quality, while the visibility (\textit{VG} \& \textit{VA}) further enhances the consistency. }
    \label{tab:ablation_results}
    \resizebox{\linewidth}{!}{
    \begin{tabular}{p{0.25cm}p{0.25cm}p{0.25cm}p{0.25cm}cccccccccccc} 
    \toprule
    \multirow{2.5}{*}{\textit{GP}} & \multirow{2.5}{*}{\textit{VG}} & \multirow{2.5}{*}{\textit{AP}} & \multirow{2.5}{*}{\textit{VA}} & \multicolumn{2}{c}{Scene Recon.}                                               & \multicolumn{2}{c}{Rendering}                                                                               & \multicolumn{2}{c}{Object Recon.}                                                                         & \multicolumn{2}{c}{BG Recon.}                                                  \\ \cmidrule(lr){5-6} \cmidrule(lr){7-8}\cmidrule(lr){9-10} \cmidrule(lr){11-12}
                       & & & & CD$\downarrow$ & NC$\uparrow$ & PSNR$\uparrow$ & MUSIQ$\uparrow$ & CD$\downarrow$ & mIoU$\uparrow$ & CD$\downarrow$ & NC$\uparrow$ \\ 
    \midrule
        $\times$ & $\times$ & $\times$ & $\times$                   & 8.51                   & 86.13                  & 24.31                   & 41.51                     & 7.67                   & 73.31                    & 10.06                  & 87.36                  \\
        $\checkmark$ & $\times$ & $\times$ & $\times$               & 8.14                   & 88.68                  & 24.83                   & 41.98                     & 6.35                   & 84.45                    & 9.83                   & 91.36                  \\
        $\checkmark$ & $\checkmark$ & $\times$ & $\times$               & 7.91                   & 89.36                  & 25.08                   & 43.33                     & 5.54                   & 88.21                    & 9.39                   & 92.83                  \\
        $\checkmark$ & $\times$ & $\checkmark$ & $\times$               & 8.14                   & 88.68                  & 22.83                   & 50.25                     & 6.35                   & 84.45                    & 9.83                   & 91.36                  \\
        $\checkmark$ & $\checkmark$ & $\checkmark$ & $\times$               & 7.91                   & 89.36                  & 23.42                   & 52.34                     & 5.54                   & 88.21                    & 9.39                   & 92.83                  \\
        $\checkmark$ & $\checkmark$ & $\checkmark$ & $\checkmark$ & 7.91                   & 89.36                  & 24.52                   & 49.22                     & 5.54                   & 88.21                    & 9.39                   & 92.83                  \\ 
        \bottomrule
    \end{tabular}
    }
\end{table}

\paragraph{Holistic Scene Reconstruction}
As shown in \cref{tab:quantitative_scene_results}, our method enhances both reconstruction and rendering results compared to all baselines. These improvements stem from incorporating generative priors into the reconstruction pipeline, which enables more accurate reconstruction in less captured areas, more precise object structures, smoother background reconstruction, and fewer floating artifacts, as illustrated in ~\cref{fig:mesh_results}. 
Our appearance prior also supplies reasonable additional information in these less captured regions, allowing our \ac{nr} rendering metric, MUSIQ, to significantly outperform the baselines. This indicates our rendering achieves higher quality in these areas, in contrast to the artifacts present in the baseline results, as shown in \cref{fig:color_results}.

\begin{figure}[t]
    \centering
    \includegraphics[width=\linewidth]{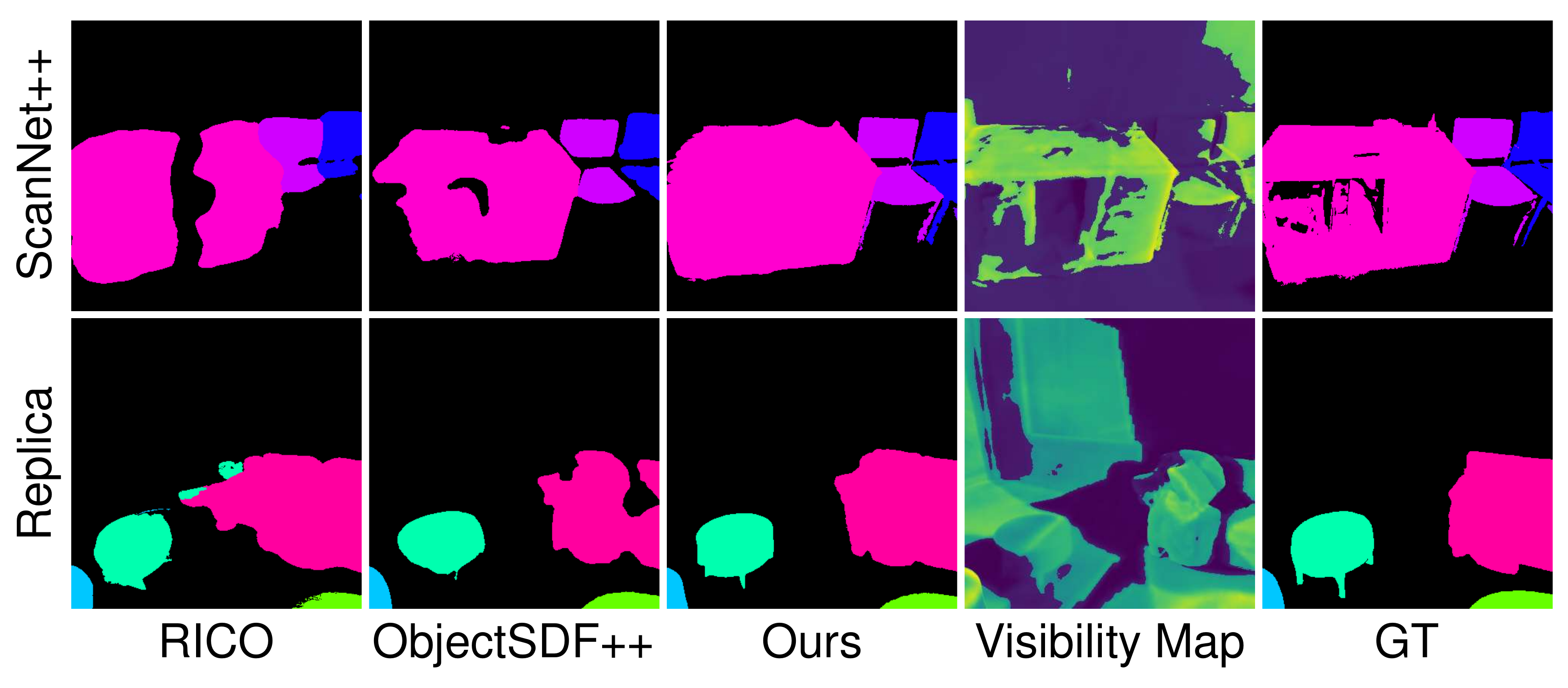}
    \caption{\textbf{Visualized novel view instance masks.} Our method can synthesize consistent and complete novel view instance masks.}
    \label{fig:mask_results}
\end{figure}

\paragraph{Decompositional Object Reconstruction}
Incorporating generative prior substantially improves the reconstruction in occluded regions, including more accurate object structures and fewer floating artifacts, \eg, the chair behind the table and the occluded background in \cref{fig:mesh_results}. The significant increase in novel view mask \ac{miou} compared to all baselines shows that our method achieves complete and multi-view consistent object shapes, as illustrated in ~\cref{tab:quantitative_object_results} and ~\cref{fig:mask_results}. Our results, shown in ~\cref{fig:mesh_results}, remain faithful to input images in observed regions, confirming that our visibility-guided approach effectively mitigates conflicts between the guidance of generative prior and the input images.

\paragraph{Performance under Different View Number}
\cref{tab:view_results} shows our result consistently outperforms the baselines across varying numbers of input views, improving both geometry and appearance. Notably, it realizes better object reconstruction with just 5 views than baselines with 15 views. Our method is especially effective in large scenes with heavy occlusions, as demonstrated in \cref{fig:teaser} where our method with 10 views outperforms the baseline with 100.

\subsection{Ablation Studies}
\begin{figure}[htbp]
    \centering
    \includegraphics[width=\linewidth]{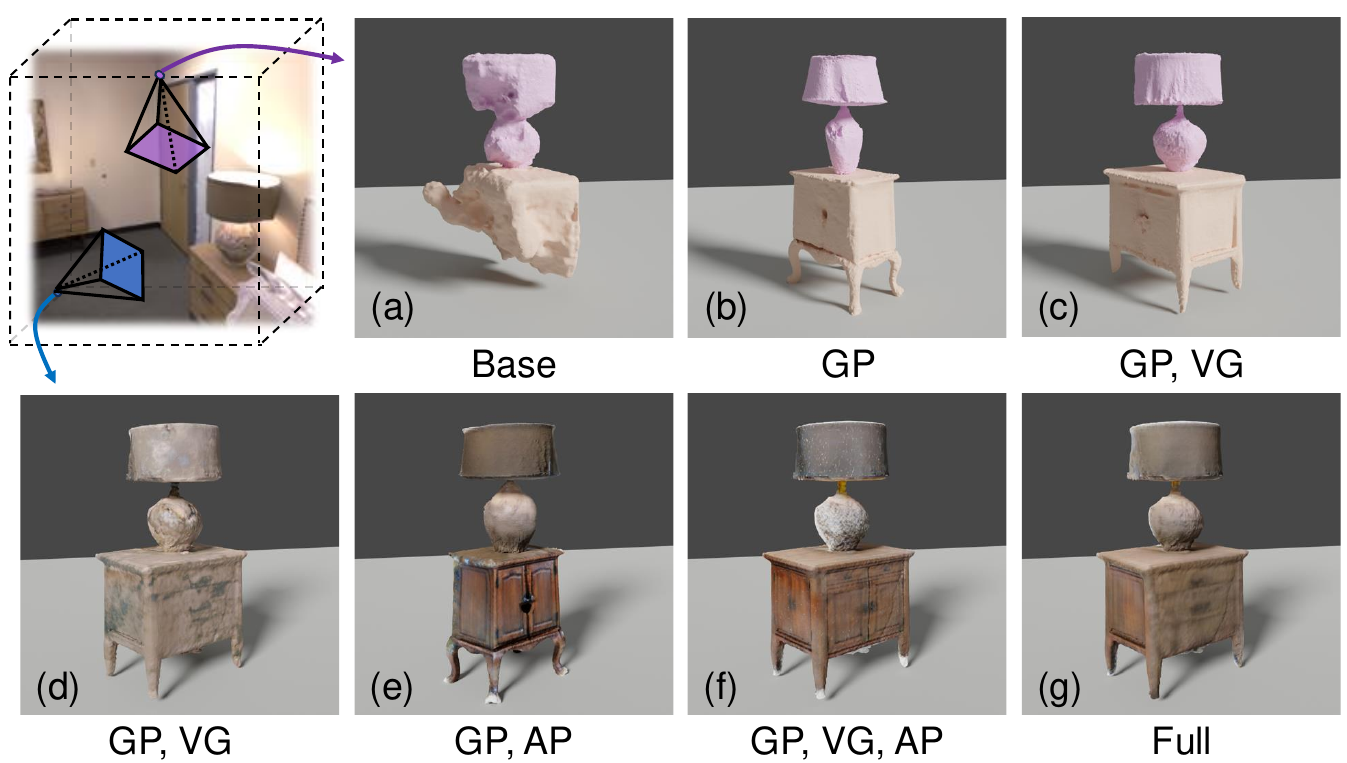}
    \vspace{-6mm}
    \caption{\textbf{Qualitative ablation comparison.} We show the meshes in the first row along with their textures in the second, demonstrating that prior knowledge can supplement missing information while the visibility modeling ensures consistency with input views.}
    \vspace{-2mm}
    \label{fig:ablation}
\end{figure}

We design ablative studies on the geometry prior (\textit{GP}), visibility guidance for geometry prior (\textit{VG}), appearance prior (\textit{AP}), and visibility guidance for appearance prior (\textit{VA}). \cref{tab:ablation_results} and \cref{fig:ablation} reveal several key observations:
\begin{enumerate}[leftmargin=*]
    \item The integration of generative priors (\textit{GP} \& \textit{AP}) substantially improves reconstruction and rendering quality. However, directly incorporating them creates inconsistencies with input views, potentially undermining the reconstruction of observed regions (see \cref{fig:ablation} (b, e, f)). 
    \item As shown in \cref{tab:ablation_results} and \cref{fig:ablation} (c, g), visibility guidance (\textit{VG} and \textit{VA}) effectively regulates the influence of diffusion priors. By adaptively weighting the \ac{sds} loss, \ie, reducing its impact in high-visibility regions and amplifying it in low-visibility areas, our method achieves an optimal balance in prior-guided reconstruction.
    \item The full model, incorporating all components, achieves the best overall performance. This demonstrates the effectiveness of our designs for geometry and appearance reconstruction, especially in unobserved regions.
\end{enumerate}

\subsection{Scene Editing}

\begin{figure}[htbp]
    \centering
    \includegraphics[width=\linewidth]{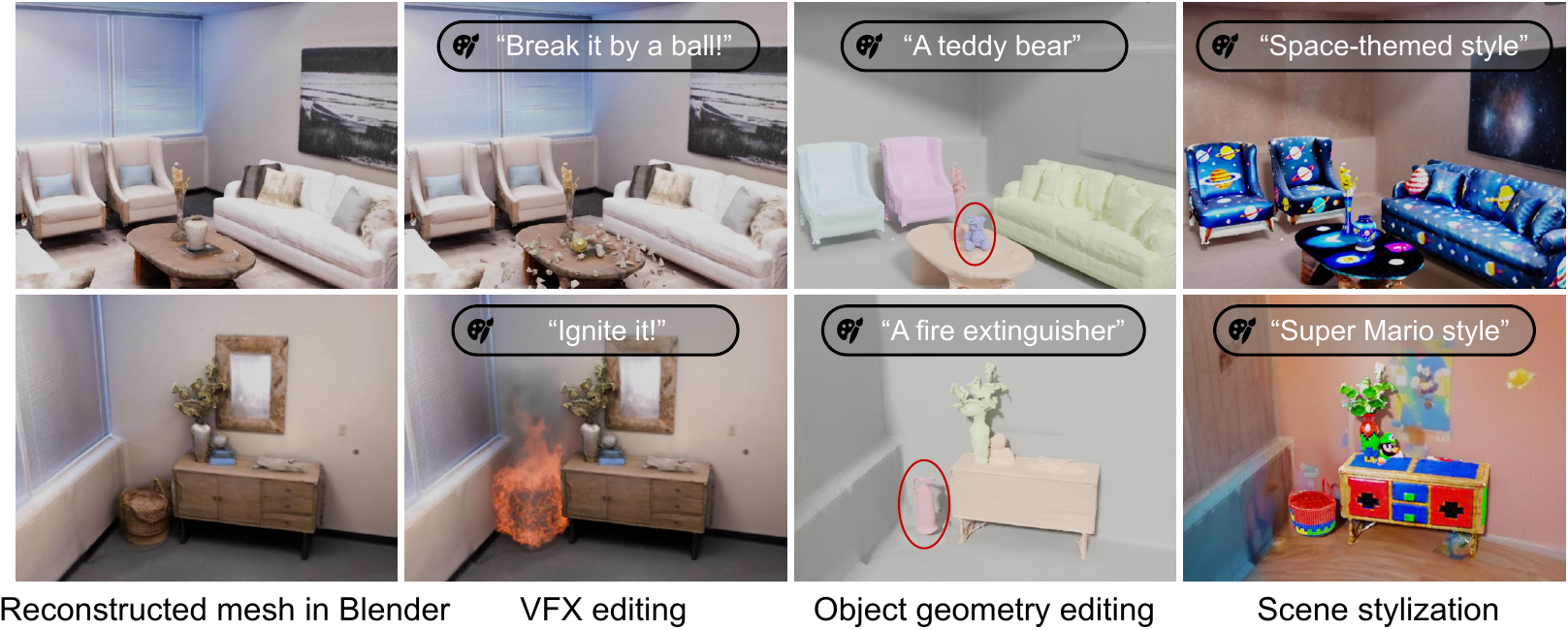}
    \caption{\textbf{Examples of scene editing}. Based on reconstructed scenes, our model seamlessly supports flexible text-guided geometry and appearance editing, as well as \ac{vfx} editing.}
    \vspace{-2mm}
    \label{fig:editing}
\end{figure}

\paragraph{Text-based Editing}
Leveraging decompositional reconstruction and text-guided generative priors, \modelname enables seamless text-based editing of both geometry (\eg, \textit{``A Teddy bear''}) and appearance (\eg, \textit{``Space-themed''}) for individual objects and background, as shown in \cref{fig:editing}. This allows the generation of numerous digital replicas~\cite{replica19arxiv} or cousins~\cite{dai2024acdc} while preserving the original spatial layout.

\paragraph{\ac{vfx} Editing}
Our method reconstructs high-fidelity, decomposed object meshes with detailed UV maps, supporting \ac{vfx} workflows in common 3D software like Blender. 
\cref{fig:editing} demonstrates diverse photorealistic \ac{vfx} edits for individual objects (\eg, \textit{``Ignite it''} or \textit{``Break it by a ball''}), which could benefit filmmaking and game development.

\section{Conclusion}
We present \modelname, a novel pipeline that utilizes generative priors in the form of \ac{sds} to optimize the neural representation of each object. It employs a visibility-guided mechanism that dynamically adjusts the \ac{sds} loss, ensuring improved results while maintaining consistency with the input images. Extensive experiments demonstrate that our method outperforms \sota methods in both geometry and appearance reconstruction, effectively enhancing quality in unobserved regions while preserving fidelity in observed areas. Our method enables seamless text-based editing for geometry and stylization, and generates decomposed object meshes with detailed UV maps, supporting a wide range of downstream applications.


\small
\bibliographystyle{ieeenat_fullname}
\bibliography{main}

\begin{thebibliography}{109}
\providecommand{\natexlab}[1]{#1}
\providecommand{\url}[1]{\texttt{#1}}
\expandafter\ifx\csname urlstyle\endcsname\relax
  \providecommand{\doi}[1]{doi: #1}\else
  \providecommand{\doi}{doi: \begingroup \urlstyle{rm}\Url}\fi

\bibitem[Anderson et~al.(2018)Anderson, Wu, Teney, Bruce, Johnson, S{\"u}nderhauf, Reid, Gould, and Van Den~Hengel]{anderson2018vision}
Peter Anderson, Qi Wu, Damien Teney, Jake Bruce, Mark Johnson, Niko S{\"u}nderhauf, Ian Reid, Stephen Gould, and Anton Van Den~Hengel.
\newblock Vision-and-language navigation: Interpreting visually-grounded navigation instructions in real environments.
\newblock In \emph{Conference on Computer Vision and Pattern Recognition (CVPR)}, 2018.

\bibitem[Andrew(2023)]{sdxlstyle}
Andrew.
\newblock Sdxl-styles.
\newblock \url{https://stable-diffusion-art.com/sdxl-styles/}, 2023.

\bibitem[Barron et~al.(2022)Barron, Mildenhall, Verbin, Srinivasan, and Hedman]{barron2022mip}
Jonathan~T Barron, Ben Mildenhall, Dor Verbin, Pratul~P Srinivasan, and Peter Hedman.
\newblock Mip-nerf 360: Unbounded anti-aliased neural radiance fields.
\newblock In \emph{Conference on Computer Vision and Pattern Recognition (CVPR)}, 2022.

\bibitem[Barron et~al.(2023)Barron, Mildenhall, Verbin, Srinivasan, and Hedman]{barron2023zip}
Jonathan~T Barron, Ben Mildenhall, Dor Verbin, Pratul~P Srinivasan, and Peter Hedman.
\newblock Zip-nerf: Anti-aliased grid-based neural radiance fields.
\newblock In \emph{International Conference on Computer Vision (ICCV)}, pages 19697--19705, 2023.

\bibitem[Chen et~al.(2024{\natexlab{a}})Chen, Loy, and Pan]{chen2024mvip}
Honghua Chen, Chen~Change Loy, and Xingang Pan.
\newblock Mvip-nerf: Multi-view 3d inpainting on nerf scenes via diffusion prior.
\newblock In \emph{Conference on Computer Vision and Pattern Recognition (CVPR)}, 2024{\natexlab{a}}.

\bibitem[Chen et~al.(2023)Chen, Chen, Jiao, and Jia]{chen2023fantasia3d}
Rui Chen, Yongwei Chen, Ningxin Jiao, and Kui Jia.
\newblock Fantasia3d: Disentangling geometry and appearance for high-quality text-to-3d content creation.
\newblock In \emph{International Conference on Computer Vision (ICCV)}, 2023.

\bibitem[Chen et~al.(2024{\natexlab{b}})Chen, Chen, Zhang, Wang, Yang, Wang, Cai, Yang, Liu, and Lin]{chen2023gaussianeditor}
Yiwen Chen, Zilong Chen, Chi Zhang, Feng Wang, Xiaofeng Yang, Yikai Wang, Zhongang Cai, Lei Yang, Huaping Liu, and Guosheng Lin.
\newblock Gaussianeditor: Swift and controllable 3d editing with gaussian splatting.
\newblock In \emph{Conference on Computer Vision and Pattern Recognition (CVPR)}, 2024{\natexlab{b}}.

\bibitem[Chen et~al.(2024{\natexlab{c}})Chen, Ni, Jiang, Zhang, Zhu, and Huang]{chen2023ssr}
Yixin Chen, Junfeng Ni, Nan Jiang, Yaowei Zhang, Yixin Zhu, and Siyuan Huang.
\newblock Single-view 3d scene reconstruction with high-fidelity shape and texture.
\newblock In \emph{International Conference on 3D Vision (3DV)}, 2024{\natexlab{c}}.

\bibitem[Chen et~al.(2024{\natexlab{d}})Chen, Wang, Wu, Pan, Jia, and Liu]{chen2024comboverse}
Yongwei Chen, Tengfei Wang, Tong Wu, Xingang Pan, Kui Jia, and Ziwei Liu.
\newblock Comboverse: Compositional 3d assets creation using spatially-aware diffusion guidance.
\newblock In \emph{European Conference on Computer Vision (ECCV)}, 2024{\natexlab{d}}.

\bibitem[Chibane et~al.(2020)Chibane, Alldieck, and Pons-Moll]{chibane2020implicit}
Julian Chibane, Thiemo Alldieck, and Gerard Pons-Moll.
\newblock Implicit functions in feature space for 3d shape reconstruction and completion.
\newblock In \emph{Conference on Computer Vision and Pattern Recognition (CVPR)}, 2020.

\bibitem[Community(2018)]{blender}
Blender~Online Community.
\newblock Blender - a 3d modelling and rendering package.
\newblock Blender Foundation, Stichting Blender Foundation, Amsterdam, 2018.

\bibitem[Cui et~al.(2023)Cui, Gong, Jia, Huang, Zheng, Ma, and Zhu]{cui2023probio}
Jieming Cui, Ziren Gong, Baoxiong Jia, Siyuan Huang, Zilong Zheng, Jianzhu Ma, and Yixin Zhu.
\newblock Probio: A protocol-guided multimodal dataset for molecular biology lab.
\newblock In \emph{Advances in Neural Information Processing Systems (NeurIPS)}, 2023.

\bibitem[Curless and Levoy(1996)]{curless1996tsdf}
Brian Curless and Marc Levoy.
\newblock A volumetric method for building complex models from range images.
\newblock In \emph{ACM SIGGRAPH / Eurographics Symposium on Computer Animation (SCA)}, 1996.

\bibitem[Dai et~al.(2024)Dai, Wong, Jiang, Wang, Gokmen, Zhang, Wu, and Fei-Fei]{dai2024acdc}
Tianyuan Dai, Josiah Wong, Yunfan Jiang, Chen Wang, Cem Gokmen, Ruohan Zhang, Jiajun Wu, and Li Fei-Fei.
\newblock Acdc: Automated creation of digital cousins for robust policy learning.
\newblock \emph{arXiv preprint arXiv:2410.07408}, 2024.

\bibitem[Deng et~al.(2022)Deng, Liu, Zhu, and Ramanan]{kangle2021dsnerf}
Kangle Deng, Andrew Liu, Jun-Yan Zhu, and Deva Ramanan.
\newblock Depth-supervised {NeRF}: Fewer views and faster training for free.
\newblock In \emph{Conference on Computer Vision and Pattern Recognition (CVPR)}, 2022.

\bibitem[Eftekhar et~al.(2021)Eftekhar, Sax, Malik, and Zamir]{eftekhar2021omnidata}
Ainaz Eftekhar, Alexander Sax, Jitendra Malik, and Amir Zamir.
\newblock Omnidata: A scalable pipeline for making multi-task mid-level vision datasets from 3d scans.
\newblock In \emph{International Conference on Computer Vision (ICCV)}, 2021.

\bibitem[Gal et~al.(2022)Gal, Alaluf, Atzmon, Patashnik, Bermano, Chechik, and Cohen-Or]{gal2022textual}
Rinon Gal, Yuval Alaluf, Yuval Atzmon, Or Patashnik, Amit~H. Bermano, Gal Chechik, and Daniel Cohen-Or.
\newblock An image is worth one word: Personalizing text-to-image generation using textual inversion.
\newblock \emph{arXiv preprint arXiv:2208.01618}, 2022.

\bibitem[Gao* et~al.(2024)Gao*, Holynski*, Henzler, Brussee, Martin-Brualla, Srinivasan, Barron, and Poole*]{gao2024cat3d}
Ruiqi Gao*, Aleksander Holynski*, Philipp Henzler, Arthur Brussee, Ricardo Martin-Brualla, Pratul~P. Srinivasan, Jonathan~T. Barron, and Ben Poole*.
\newblock Cat3d: Create anything in 3d with multi-view diffusion models.
\newblock In \emph{Advances in Neural Information Processing Systems (NeurIPS)}, 2024.

\bibitem[Gong et~al.(2023)Gong, Huang, Zhao, Geng, Gao, Wu, Ai, Zhou, Terzopoulos, Zhu, et~al.]{gong2023arnold}
Ran Gong, Jiangyong Huang, Yizhou Zhao, Haoran Geng, Xiaofeng Gao, Qingyang Wu, Wensi Ai, Ziheng Zhou, Demetri Terzopoulos, Song-Chun Zhu, et~al.
\newblock Arnold: A benchmark for language-grounded task learning with continuous states in realistic 3d scenes.
\newblock \emph{arXiv preprint arXiv:2304.04321}, 2023.

\bibitem[Gu et~al.(2023)Gu, Xiang, Li, Ling, Liu, Mu, Tang, Tao, Wei, Yao, et~al.]{gu2023maniskill2}
Jiayuan Gu, Fanbo Xiang, Xuanlin Li, Zhan Ling, Xiqiang Liu, Tongzhou Mu, Yihe Tang, Stone Tao, Xinyue Wei, Yunchao Yao, et~al.
\newblock Maniskill2: A unified benchmark for generalizable manipulation skills.
\newblock \emph{arXiv preprint arXiv:2302.04659}, 2023.

\bibitem[Gui et~al.(2025)Gui, Fischer, Prestel, Ma, Kotovenko, Grebenkova, Baumann, Hu, and Ommer]{gui2024depthfm}
Ming Gui, Johannes~S. Fischer, Ulrich Prestel, Pingchuan Ma, Dmytro Kotovenko, Olga Grebenkova, Stefan~Andreas Baumann, Vincent~Tao Hu, and Björn Ommer.
\newblock Depthfm: Fast monocular depth estimation with flow matching.
\newblock In \emph{AAAI Conference on Artificial Intelligence (AAAI)}, 2025.

\bibitem[Hu et~al.(2023)Hu, Zhou, Li, Yu, Hong, Hu, Li, Lee, and Liu]{hu2023consistentnerf}
Shoukang Hu, Kaichen Zhou, Kaiyu Li, Longhui Yu, Lanqing Hong, Tianyang Hu, Zhenguo Li, Gim~Hee Lee, and Ziwei Liu.
\newblock Consistentnerf: Enhancing neural radiance fields with 3d consistency for sparse view synthesis.
\newblock \emph{arXiv preprint arXiv:2305.11031}, 2023.

\bibitem[Huang et~al.(2024)Huang, Yong, Ma, Linghu, Li, Wang, Li, Zhu, Jia, and Huang]{huang2023embodied}
Jiangyong Huang, Silong Yong, Xiaojian Ma, Xiongkun Linghu, Puhao Li, Yan Wang, Qing Li, Song-Chun Zhu, Baoxiong Jia, and Siyuan Huang.
\newblock An embodied generalist agent in 3d world.
\newblock In \emph{International Conference on Machine Learning (ICML)}, 2024.

\bibitem[Huang et~al.(2025)Huang, Jia, Wang, Zhu, Linghu, Li, Zhu, and Huang]{huang2025unveiling}
Jiangyong Huang, Baoxiong Jia, Yan Wang, Ziyu Zhu, Xiongkun Linghu, Qing Li, Song-Chun Zhu, and Siyuan Huang.
\newblock Unveiling the mist over 3d vision-language understanding: Object-centric evaluation with chain-of-analysis.
\newblock In \emph{Conference on Computer Vision and Pattern Recognition (CVPR)}, 2025.

\bibitem[Jain et~al.(2021)Jain, Tancik, and Abbeel]{jain2021putting}
Ajay Jain, Matthew Tancik, and Pieter Abbeel.
\newblock Putting nerf on a diet: Semantically consistent few-shot view synthesis.
\newblock In \emph{International Conference on Computer Vision (ICCV)}, 2021.

\bibitem[Jiang et~al.(2024{\natexlab{a}})Jiang, He, Wang, Li, Chen, Huang, and Zhu]{jiang2024autonomous}
Nan Jiang, Zimo He, Zi Wang, Hongjie Li, Yixin Chen, Siyuan Huang, and Yixin Zhu.
\newblock Autonomous character-scene interaction synthesis from text instruction.
\newblock In \emph{SIGGRAPH Asia 2024 Conference Papers}, 2024{\natexlab{a}}.

\bibitem[Jiang et~al.(2024{\natexlab{b}})Jiang, Zhang, Li, Ma, Wang, Chen, Liu, Zhu, and Huang]{jiang2024scaling}
Nan Jiang, Zhiyuan Zhang, Hongjie Li, Xiaoxuan Ma, Zan Wang, Yixin Chen, Tengyu Liu, Yixin Zhu, and Siyuan Huang.
\newblock Scaling up dynamic human-scene interaction modeling.
\newblock In \emph{Conference on Computer Vision and Pattern Recognition (CVPR)}, 2024{\natexlab{b}}.

\bibitem[Jiang et~al.(2022)Jiang, Gupta, Zhang, Wang, Dou, Chen, Fei-Fei, Anandkumar, Zhu, and Fan]{jiang2022vima}
Yunfan Jiang, Agrim Gupta, Zichen Zhang, Guanzhi Wang, Yongqiang Dou, Yanjun Chen, Li Fei-Fei, Anima Anandkumar, Yuke Zhu, and Linxi Fan.
\newblock Vima: General robot manipulation with multimodal prompts.
\newblock \emph{arXiv preprint arXiv:2210.03094}, 2022.

\bibitem[Ke et~al.(2021)Ke, Wang, Wang, Milanfar, and Yang]{ke2021musiq}
Junjie Ke, Qifei Wang, Yilin Wang, Peyman Milanfar, and Feng Yang.
\newblock Musiq: Multi-scale image quality transformer.
\newblock In \emph{International Conference on Computer Vision (ICCV)}, 2021.

\bibitem[Kerbl et~al.(2023)Kerbl, Kopanas, Leimk{\"u}hler, and Drettakis]{kerbl3Dgaussians}
Bernhard Kerbl, Georgios Kopanas, Thomas Leimk{\"u}hler, and George Drettakis.
\newblock 3d gaussian splatting for real-time radiance field rendering.
\newblock In \emph{ACM SIGGRAPH / Eurographics Symposium on Computer Animation (SCA)}, 2023.

\bibitem[Kim et~al.(2022)Kim, Seo, and Han]{kim2022infonerf}
Mijeong Kim, Seonguk Seo, and Bohyung Han.
\newblock Infonerf: Ray entropy minimization for few-shot neural volume rendering.
\newblock In \emph{Conference on Computer Vision and Pattern Recognition (CVPR)}, 2022.

\bibitem[Kingma and Ba(2014)]{kingma2014adam}
Diederik~P Kingma and Jimmy Ba.
\newblock Adam: A method for stochastic optimization.
\newblock \emph{arXiv preprint arXiv:1412.6980}, 2014.

\bibitem[Kirillov et~al.(2023)Kirillov, Mintun, Ravi, Mao, Rolland, Gustafson, Xiao, Whitehead, Berg, Lo, Doll{\'a}r, and Girshick]{kirillov2023segany}
Alexander Kirillov, Eric Mintun, Nikhila Ravi, Hanzi Mao, Chloe Rolland, Laura Gustafson, Tete Xiao, Spencer Whitehead, Alexander~C. Berg, Wan-Yen Lo, Piotr Doll{\'a}r, and Ross Girshick.
\newblock Segment anything.
\newblock In \emph{International Conference on Computer Vision (ICCV)}, 2023.

\bibitem[Kong et~al.(2023)Kong, Liu, Taher, and Davison]{kong2023vmap}
Xin Kong, Shikun Liu, Marwan Taher, and Andrew~J Davison.
\newblock vmap: Vectorised object mapping for neural field slam.
\newblock In \emph{Conference on Computer Vision and Pattern Recognition (CVPR)}, 2023.

\bibitem[Krantz et~al.(2020)Krantz, Wijmans, Majumdar, Batra, and Lee]{krantz2020beyond}
Jacob Krantz, Erik Wijmans, Arjun Majumdar, Dhruv Batra, and Stefan Lee.
\newblock Beyond the nav-graph: Vision-and-language navigation in continuous environments.
\newblock In \emph{European Conference on Computer Vision (ECCV)}, 2020.

\bibitem[Kwak et~al.(2023)Kwak, Song, and Kim]{kwak2023geconerf}
Minseop Kwak, Jiuhn Song, and Seungryong Kim.
\newblock Geconerf: Few-shot neural radiance fields via geometric consistency.
\newblock In \emph{International Conference on Machine Learning (ICML)}, 2023.

\bibitem[Laine et~al.(2020)Laine, Hellsten, Karras, Seol, Lehtinen, and Aila]{Laine2020diffrast}
Samuli Laine, Janne Hellsten, Tero Karras, Yeongho Seol, Jaakko Lehtinen, and Timo Aila.
\newblock Modular primitives for high-performance differentiable rendering.
\newblock In \emph{ACM SIGGRAPH / Eurographics Symposium on Computer Animation (SCA)}, 2020.

\bibitem[Lee et~al.(2024)Lee, Sohn, and Shin]{lee2024dreamflow}
Kyungmin Lee, Kihyuk Sohn, and Jinwoo Shin.
\newblock Dreamflow: High-quality text-to-3d generation by approximating probability flow.
\newblock \emph{arXiv preprint arXiv:2403.14966}, 2024.

\bibitem[Li et~al.(2024{\natexlab{a}})Li, Liu, Li, Han, Geng, Wang, Zhu, Zhu, and Huang]{li2024ag2manip}
Puhao Li, Tengyu Liu, Yuyang Li, Muzhi Han, Haoran Geng, Shu Wang, Yixin Zhu, Song-Chun Zhu, and Siyuan Huang.
\newblock Ag2manip: Learning novel manipulation skills with agent-agnostic visual and action representations.
\newblock In \emph{International Conference on Intelligent Robots and Systems (IROS)}, 2024{\natexlab{a}}.

\bibitem[Li et~al.(2023{\natexlab{a}})Li, Gao, Tancik, and Kanazawa]{li2023nerfacc}
Ruilong Li, Hang Gao, Matthew Tancik, and Angjoo Kanazawa.
\newblock Nerfacc: Efficient sampling accelerates nerfs.
\newblock In \emph{International Conference on Computer Vision (ICCV)}, 2023{\natexlab{a}}.

\bibitem[Li et~al.(2024{\natexlab{b}})Li, Liu, Geng, Li, Yang, Zhu, Liu, and Huang]{li2024grasp}
Yuyang Li, Bo Liu, Yiran Geng, Puhao Li, Yaodong Yang, Yixin Zhu, Tengyu Liu, and Siyuan Huang.
\newblock Grasp multiple objects with one hand.
\newblock In \emph{International Conference on Intelligent Robots and Systems (IROS)}, 2024{\natexlab{b}}.

\bibitem[Li et~al.(2023{\natexlab{b}})Li, Lyu, Ding, Wang, Liao, and Liu]{li2023rico}
Zizhang Li, Xiaoyang Lyu, Yuanyuan Ding, Mengmeng Wang, Yiyi Liao, and Yong Liu.
\newblock Rico: Regularizing the unobservable for indoor compositional reconstruction.
\newblock In \emph{International Conference on Computer Vision (ICCV)}, 2023{\natexlab{b}}.

\bibitem[Liu et~al.(2024{\natexlab{a}})Liu, Sun, Wang, Wang, Sun, Ye, Zhang, and Duan]{liu2024reconx}
Fangfu Liu, Wenqiang Sun, Hanyang Wang, Yikai Wang, Haowen Sun, Junliang Ye, Jun Zhang, and Yueqi Duan.
\newblock Reconx: Reconstruct any scene from sparse views with video diffusion model.
\newblock \emph{arXiv preprint arXiv:2408.16767}, 2024{\natexlab{a}}.

\bibitem[Liu et~al.(2022)Liu, Zheng, Chen, Cui, and Han]{liu2022towards}
Haolin Liu, Yujian Zheng, Guanying Chen, Shuguang Cui, and Xiaoguang Han.
\newblock Towards high-fidelity single-view holistic reconstruction of indoor scenes.
\newblock In \emph{European Conference on Computer Vision (ECCV)}, 2022.

\bibitem[Liu et~al.(2023)Liu, Wu, Hoorick, Tokmakov, Zakharov, and Vondrick]{liu2023zero1to3}
Ruoshi Liu, Rundi Wu, Basile~Van Hoorick, Pavel Tokmakov, Sergey Zakharov, and Carl Vondrick.
\newblock Zero-1-to-3: Zero-shot one image to 3d object.
\newblock In \emph{International Conference on Computer Vision (ICCV)}, 2023.

\bibitem[Liu et~al.(2024{\natexlab{b}})Liu, Zhou, and Huang]{liu20243dgsenhancer}
Xi Liu, Chaoyi Zhou, and Siyu Huang.
\newblock 3dgs-enhancer: Enhancing unbounded 3d gaussian splatting with view-consistent 2d diffusion priors.
\newblock \emph{arXiv preprint arXiv:2410.16266}, 2024{\natexlab{b}}.

\bibitem[Liu et~al.(2024{\natexlab{c}})Liu, Jia, Chen, and Huang]{Liu2024slotlifter}
Yu Liu, Baoxiong Jia, Yixin Chen, and Siyuan Huang.
\newblock Slotlifter: Slot-guided feature lifting for learning object-centric radiance fields.
\newblock In \emph{European Conference on Computer Vision (ECCV)}, 2024{\natexlab{c}}.

\bibitem[Liu et~al.(2025)Liu, Jia, Lu, Ni, Zhu, and Huang]{liu2025building}
Yu Liu, Baoxiong Jia, Ruijie Lu, Junfeng Ni, Song-Chun Zhu, and Siyuan Huang.
\newblock Building interactable replicas of complex articulated objects via gaussian splatting.
\newblock In \emph{International Conference on Learning Representations (ICLR)}, 2025.

\bibitem[Lu et~al.(2024)Lu, Zhang, Wang, Liu, Lu, and Tang]{lu2024manigaussian}
Guanxing Lu, Shiyi Zhang, Ziwei Wang, Changliu Liu, Jiwen Lu, and Yansong Tang.
\newblock Manigaussian: Dynamic gaussian splatting for multi-task robotic manipulation.
\newblock In \emph{European Conference on Computer Vision (ECCV)}, 2024.

\bibitem[Lu et~al.(2025{\natexlab{a}})Lu, Chen, Liu, Tang, Ni, Wan, Zeng, and Huang]{lu2025taco}
Ruijie Lu, Yixin Chen, Yu Liu, Jiaxiang Tang, Junfeng Ni, Diwen Wan, Gang Zeng, and Siyuan Huang.
\newblock Taco: Taming diffusion for in-the-wild video amodal completion.
\newblock \emph{arXiv preprint arXiv:2503.12049}, 2025{\natexlab{a}}.

\bibitem[Lu et~al.(2025{\natexlab{b}})Lu, Chen, Ni, Jia, Liu, Wan, Zeng, and Huang]{lu2024movis}
Ruijie Lu, Yixin Chen, Junfeng Ni, Baoxiong Jia, Yu Liu, Diwen Wan, Gang Zeng, and Siyuan Huang.
\newblock Movis: Enhancing multi-object novel view synthesis for indoor scenes.
\newblock In \emph{Conference on Computer Vision and Pattern Recognition (CVPR)}, 2025{\natexlab{b}}.

\bibitem[Lyu et~al.(2024)Lyu, Chang, Dai, Sun, and Qi]{lyu2024total}
Xiaoyang Lyu, Chirui Chang, Peng Dai, Yang-Tian Sun, and Xiaojuan Qi.
\newblock Total-decom: Decomposed 3d scene reconstruction with minimal interaction.
\newblock In \emph{Conference on Computer Vision and Pattern Recognition (CVPR)}, 2024.

\bibitem[Martel et~al.(2021)Martel, Lindell, Lin, Chan, Monteiro, and Wetzstein]{martel2021acorn}
Julien~NP Martel, David~B Lindell, Connor~Z Lin, Eric~R Chan, Marco Monteiro, and Gordon Wetzstein.
\newblock Acorn: Adaptive coordinate networks for neural scene representation.
\newblock \emph{ACM SIGGRAPH / Eurographics Symposium on Computer Animation (SCA)}, 2021.

\bibitem[McAllister et~al.(2024)McAllister, Ge, Huang, Jacobs, Efros, Holynski, and Kanazawa]{mcallister2024rethinking}
David McAllister, Songwei Ge, Jia-Bin Huang, David~W Jacobs, Alexei~A Efros, Aleksander Holynski, and Angjoo Kanazawa.
\newblock Rethinking score distillation as a bridge between image distributions.
\newblock \emph{arXiv preprint arXiv:2406.09417}, 2024.

\bibitem[Melas-Kyriazi et~al.(2024)Melas-Kyriazi, Laina, Rupprecht, Neverova, Vedaldi, Gafni, and Kokkinos]{melaskyriazi2024im3d}
Luke Melas-Kyriazi, Iro Laina, Christian Rupprecht, Natalia Neverova, Andrea Vedaldi, Oran Gafni, and Filippos Kokkinos.
\newblock Im-3d: Iterative multiview diffusion and reconstruction for high-quality 3d generation.
\newblock In \emph{International Conference on Machine Learning (ICML)}, 2024.

\bibitem[Mescheder et~al.(2019)Mescheder, Oechsle, Niemeyer, Nowozin, and Geiger]{mescheder2019occupancy}
Lars Mescheder, Michael Oechsle, Michael Niemeyer, Sebastian Nowozin, and Andreas Geiger.
\newblock Occupancy networks: Learning 3d reconstruction in function space.
\newblock In \emph{Conference on Computer Vision and Pattern Recognition (CVPR)}, 2019.

\bibitem[Mildenhall et~al.(2020)Mildenhall, Srinivasan, Tancik, Barron, Ramamoorthi, and Ng]{mildenhall2020nerf}
Ben Mildenhall, Pratul~P. Srinivasan, Matthew Tancik, Jonathan~T. Barron, Ravi Ramamoorthi, and Ren Ng.
\newblock Nerf: Representing scenes as neural radiance fields for view synthesis.
\newblock In \emph{European Conference on Computer Vision (ECCV)}, 2020.

\bibitem[Ni et~al.(2024)Ni, Chen, Jing, Jiang, Wang, Dai, Li, Zhu, Zhu, and Huang]{ni2024phyrecon}
Junfeng Ni, Yixin Chen, Bohan Jing, Nan Jiang, Bin Wang, Bo Dai, Puhao Li, Yixin Zhu, Song-Chun Zhu, and Siyuan Huang.
\newblock Phyrecon: Physically plausible neural scene reconstruction.
\newblock In \emph{Advances in Neural Information Processing Systems (NeurIPS)}, 2024.

\bibitem[Nie et~al.(2020)Nie, Han, Guo, Zheng, Chang, and Zhang]{nie2020cvpr}
Yinyu Nie, Xiaoguang Han, Shihui Guo, Yujian Zheng, Jian Chang, and Jian~Jun Zhang.
\newblock Total3dunderstanding: Joint layout, object pose and mesh reconstruction for indoor scenes from a single image.
\newblock In \emph{Conference on Computer Vision and Pattern Recognition (CVPR)}, 2020.

\bibitem[Niemeyer et~al.(2020)Niemeyer, Mescheder, Oechsle, and Geiger]{niemeyer2020differentiable}
Michael Niemeyer, Lars Mescheder, Michael Oechsle, and Andreas Geiger.
\newblock Differentiable volumetric rendering: Learning implicit 3d representations without 3d supervision.
\newblock In \emph{Conference on Computer Vision and Pattern Recognition (CVPR)}, 2020.

\bibitem[Niemeyer et~al.(2022)Niemeyer, Barron, Mildenhall, Sajjadi, Geiger, and Radwan]{niemeyer2021regnerf}
Michael Niemeyer, Jonathan~T. Barron, Ben Mildenhall, Mehdi S.~M. Sajjadi, Andreas Geiger, and Noha Radwan.
\newblock Regnerf: Regularizing neural radiance fields for view synthesis from sparse inputs.
\newblock In \emph{Conference on Computer Vision and Pattern Recognition (CVPR)}, 2022.

\bibitem[Oechsle et~al.(2021)Oechsle, Peng, and Geiger]{oechsle2021unisurf}
Michael Oechsle, Songyou Peng, and Andreas Geiger.
\newblock Unisurf: Unifying neural implicit surfaces and radiance fields for multi-view reconstruction.
\newblock In \emph{International Conference on Computer Vision (ICCV)}, 2021.

\bibitem[Park et~al.(2019)Park, Florence, Straub, Newcombe, and Lovegrove]{park2019deepsdf}
Jeong~Joon Park, Peter Florence, Julian Straub, Richard Newcombe, and Steven Lovegrove.
\newblock Deepsdf: Learning continuous signed distance functions for shape representation.
\newblock In \emph{Conference on Computer Vision and Pattern Recognition (CVPR)}, 2019.

\bibitem[Paszke et~al.(2019)Paszke, Gross, Massa, Lerer, Bradbury, Chanan, Killeen, Lin, Gimelshein, Antiga, et~al.]{paszke2019pytorch}
Adam Paszke, Sam Gross, Francisco Massa, Adam Lerer, James Bradbury, Gregory Chanan, Trevor Killeen, Zeming Lin, Natalia Gimelshein, Luca Antiga, et~al.
\newblock Pytorch: An imperative style, high-performance deep learning library.
\newblock In \emph{Advances in Neural Information Processing Systems (NeurIPS)}, 2019.

\bibitem[Poole et~al.(2022)Poole, Jain, Barron, and Mildenhall]{poole2022dreamfusion}
Ben Poole, Ajay Jain, Jonathan~T. Barron, and Ben Mildenhall.
\newblock Dreamfusion: Text-to-3d using 2d diffusion.
\newblock In \emph{International Conference on Learning Representations (ICLR)}, 2022.

\bibitem[Qiu et~al.(2024)Qiu, Chen, Gu, Zuo, Xu, Wu, Yuan, Dong, Bo, and Han]{qiu2024richdreamer}
Lingteng Qiu, Guanying Chen, Xiaodong Gu, Qi Zuo, Mutian Xu, Yushuang Wu, Weihao Yuan, Zilong Dong, Liefeng Bo, and Xiaoguang Han.
\newblock Richdreamer: A generalizable normal-depth diffusion model for detail richness in text-to-3d.
\newblock In \emph{Conference on Computer Vision and Pattern Recognition (CVPR)}, 2024.

\bibitem[Ravi et~al.(2024)Ravi, Gabeur, Hu, Hu, Ryali, Ma, Khedr, R{\"a}dle, Rolland, Gustafson, Mintun, Pan, Alwala, Carion, Wu, Girshick, Doll{\'a}r, and Feichtenhofer]{ravi2024sam2}
Nikhila Ravi, Valentin Gabeur, Yuan-Ting Hu, Ronghang Hu, Chaitanya Ryali, Tengyu Ma, Haitham Khedr, Roman R{\"a}dle, Chloe Rolland, Laura Gustafson, Eric Mintun, Junting Pan, Kalyan~Vasudev Alwala, Nicolas Carion, Chao-Yuan Wu, Ross Girshick, Piotr Doll{\'a}r, and Christoph Feichtenhofer.
\newblock Sam 2: Segment anything in images and videos.
\newblock \emph{arXiv preprint arXiv:2408.00714}, 2024.

\bibitem[Roessle et~al.(2022)Roessle, Barron, Mildenhall, Srinivasan, and Nie{\ss}ner]{roessle2022depthpriorsnerf}
Barbara Roessle, Jonathan~T. Barron, Ben Mildenhall, Pratul~P. Srinivasan, and Matthias Nie{\ss}ner.
\newblock Dense depth priors for neural radiance fields from sparse input views.
\newblock In \emph{Conference on Computer Vision and Pattern Recognition (CVPR)}, 2022.

\bibitem[Rombach et~al.(2022)Rombach, Blattmann, Lorenz, Esser, and Ommer]{rombach2022high}
Robin Rombach, Andreas Blattmann, Dominik Lorenz, Patrick Esser, and Bj{\"o}rn Ommer.
\newblock High-resolution image synthesis with latent diffusion models.
\newblock In \emph{Conference on Computer Vision and Pattern Recognition (CVPR)}, 2022.

\bibitem[Saharia et~al.(2022)Saharia, Chan, Saxena, Li, Whang, Denton, Ghasemipour, Ayan, Mahdavi, Lopes, Salimans, Ho, Fleet, and Norouzi]{saharia2022imagen}
Chitwan Saharia, William Chan, Saurabh Saxena, Lala Li, Jay Whang, Emily Denton, Seyed Kamyar~Seyed Ghasemipour, Burcu~Karagol Ayan, S.~Sara Mahdavi, Rapha~Gontijo Lopes, Tim Salimans, Jonathan Ho, David~J Fleet, and Mohammad Norouzi.
\newblock Photorealistic text-to-image diffusion models with deep language understanding.
\newblock In \emph{Advances in Neural Information Processing Systems (NeurIPS)}, 2022.

\bibitem[Sargent et~al.(2024)Sargent, Li, Shah, Herrmann, Yu, Zhang, Chan, Lagun, Fei-Fei, Sun, and Wu]{Kyle2024zeronvs}
Kyle Sargent, Zizhang Li, Tanmay Shah, Charles Herrmann, Hong-Xing Yu, Yunzhi Zhang, Eric~Ryan Chan, Dmitry Lagun, Li Fei-Fei, Deqing Sun, and Jiajun Wu.
\newblock {ZeroNVS}: Zero-shot 360-degree view synthesis from a single real image.
\newblock In \emph{Conference on Computer Vision and Pattern Recognition (CVPR)}, 2024.

\bibitem[Sch\"{o}nberger and Frahm(2016)]{schoenberger2016sfm}
Johannes~Lutz Sch\"{o}nberger and Jan-Michael Frahm.
\newblock Structure-from-motion revisited.
\newblock In \emph{Conference on Computer Vision and Pattern Recognition (CVPR)}, 2016.

\bibitem[Seo et~al.(2023)Seo, Han, Chang, and Kwak]{seo2023cvpr}
Seunghyeon Seo, Donghoon Han, Yeonjin Chang, and Nojun Kwak.
\newblock Mixnerf: Modeling a ray with mixture density for novel view synthesis from sparse inputs.
\newblock In \emph{Conference on Computer Vision and Pattern Recognition (CVPR)}, 2023.

\bibitem[Shi et~al.(2023)Shi, Wang, Ye, Mai, Li, and Yang]{shi2023MVDream}
Yichun Shi, Peng Wang, Jianglong Ye, Long Mai, Kejie Li, and Xiao Yang.
\newblock Mvdream: Multi-view diffusion for 3d generation.
\newblock \emph{arXiv preprint arXiv:2308.16512}, 2023.

\bibitem[Shih et~al.(2024)Shih, Ma, Boyice, Holynski, Cole, Curless, and Kontkanen]{shih2024extranerf}
Meng-Li Shih, Wei-Chiu Ma, Lorenzo Boyice, Aleksander Holynski, Forrester Cole, Brian~L. Curless, and Janne Kontkanen.
\newblock Extranerf: Visibility-aware view extrapolation of neural radiance fields with diffusion models.
\newblock In \emph{Conference on Computer Vision and Pattern Recognition (CVPR)}, 2024.

\bibitem[Shridhar et~al.(2022)Shridhar, Manuelli, and Fox]{shridhar2022cliport}
Mohit Shridhar, Lucas Manuelli, and Dieter Fox.
\newblock Cliport: What and where pathways for robotic manipulation.
\newblock In \emph{Conference on Robot Learning}, pages 894--906. PMLR, 2022.

\bibitem[Slavcheva et~al.(2024)Slavcheva, Gausebeck, Chen, Buchhofer, Sabik, Ma, Dhillon, Brandt, and Dolhasz]{matterport2024defurnishing}
Mira Slavcheva, Dave Gausebeck, Kevin Chen, David Buchhofer, Azwad Sabik, Chen Ma, Sachal Dhillon, Olaf Brandt, and Alan Dolhasz.
\newblock An empty room is all we want: Automatic defurnishing of indoor panoramas.
\newblock In \emph{IEEE/CVF Conference on Computer Vision and Pattern Recognition Workshops (CVPRW)}, 2024.

\bibitem[Somraj and Soundararajan(2023)]{somraj2023vipnerf}
Nagabhushan Somraj and Rajiv Soundararajan.
\newblock {ViP-NeRF}: Visibility prior for sparse input neural radiance fields.
\newblock In \emph{ACM SIGGRAPH / Eurographics Symposium on Computer Animation (SCA)}, 2023.

\bibitem[Somraj et~al.(2023)Somraj, Karanayil, and Soundararajan]{somraj2023simplenerf}
Nagabhushan Somraj, Adithyan Karanayil, and Rajiv Soundararajan.
\newblock {SimpleNeRF}: Regularizing sparse input neural radiance fields with simpler solutions.
\newblock In \emph{SIGGRAPH Asia}, 2023.

\bibitem[Straub et~al.(2019)Straub, Whelan, Ma, Chen, Wijmans, Green, Engel, Mur-Artal, Ren, Verma, Clarkson, Yan, Budge, Yan, Pan, Yon, Zou, Leon, Carter, Briales, Gillingham, Mueggler, Pesqueira, Savva, Batra, Strasdat, Nardi, Goesele, Lovegrove, and Newcombe]{replica19arxiv}
Julian Straub, Thomas Whelan, Lingni Ma, Yufan Chen, Erik Wijmans, Simon Green, Jakob~J. Engel, Raul Mur-Artal, Carl Ren, Shobhit Verma, Anton Clarkson, Mingfei Yan, Brian Budge, Yajie Yan, Xiaqing Pan, June Yon, Yuyang Zou, Kimberly Leon, Nigel Carter, Jesus Briales, Tyler Gillingham, Elias Mueggler, Luis Pesqueira, Manolis Savva, Dhruv Batra, Hauke~M. Strasdat, Renzo~De Nardi, Michael Goesele, Steven Lovegrove, and Richard Newcombe.
\newblock The {R}eplica dataset: A digital replica of indoor spaces.
\newblock \emph{arXiv preprint arXiv:1906.05797}, 2019.

\bibitem[Sun et~al.(2024)Sun, Xu, Yang, Zang, and Wang]{sun2024p2nerf}
Xiaotian Sun, Qingshan Xu, Xinjie Yang, Yu Zang, and Cheng Wang.
\newblock Global and hierarchical geometry consistency priors for few-shot nerfs in indoor scenes.
\newblock In \emph{Conference on Computer Vision and Pattern Recognition (CVPR)}, 2024.

\bibitem[Tang et~al.(2024)Tang, Lu, Chen, Wen, Zeng, and Liu]{tang2024intex}
Jiaxiang Tang, Ruijie Lu, Xiaokang Chen, Xiang Wen, Gang Zeng, and Ziwei Liu.
\newblock Intex: Interactive text-to-texture synthesis via unified depth-aware inpainting.
\newblock \emph{arXiv preprint arXiv:2403.11878}, 2024.

\bibitem[Truong et~al.(2023)Truong, Rakotosaona, Manhardt, and Tombari]{sparf2023}
Prune Truong, Marie-Julie Rakotosaona, Fabian Manhardt, and Federico Tombari.
\newblock Sparf: Neural radiance fields from sparse and noisy poses.
\newblock In \emph{Conference on Computer Vision and Pattern Recognition (CVPR)}, 2023.

\bibitem[Wang et~al.(2023)Wang, Chen, Loy, and Liu]{wang2022sparsenerf}
Guangcong Wang, Zhaoxi Chen, Chen~Change Loy, and Ziwei Liu.
\newblock Sparsenerf: Distilling depth ranking for few-shot novel view synthesis.
\newblock In \emph{International Conference on Computer Vision (ICCV)}, 2023.

\bibitem[Wang et~al.(2021)Wang, Liu, Liu, Theobalt, Komura, and Wang]{wang2021neus}
Peng Wang, Lingjie Liu, Yuan Liu, Christian Theobalt, Taku Komura, and Wenping Wang.
\newblock Neus: Learning neural implicit surfaces by volume rendering for multi-view reconstruction.
\newblock In \emph{Advances in Neural Information Processing Systems (NeurIPS)}, 2021.

\bibitem[Wang et~al.(2024{\natexlab{a}})Wang, Lu, Xu, Wang, Wang, Dai, Zeng, and Xu]{wang2024roomtex}
Qi Wang, Ruijie Lu, Xudong Xu, Jingbo Wang, Michael~Yu Wang, Bo Dai, Gang Zeng, and Dan Xu.
\newblock Roomtex: Texturing compositional indoor scenes via iterative inpainting.
\newblock In \emph{European Conference on Computer Vision (ECCV)}, 2024{\natexlab{a}}.

\bibitem[Wang et~al.(2024{\natexlab{b}})Wang, Lu, Wang, Bao, Li, Su, and Zhu]{wang2024prolificdreamer}
Zhengyi Wang, Cheng Lu, Yikai Wang, Fan Bao, Chongxuan Li, Hang Su, and Jun Zhu.
\newblock Prolificdreamer: High-fidelity and diverse text-to-3d generation with variational score distillation.
\newblock In \emph{Advances in Neural Information Processing Systems (NeurIPS)}, 2024{\natexlab{b}}.

\bibitem[Warburg* et~al.(2023)Warburg*, Weber*, Tancik, Hołyński, and Kanazawa]{nerfbusters2023}
Frederik Warburg*, Ethan Weber*, Matthew Tancik, Aleksander Hołyński, and Angjoo Kanazawa.
\newblock Nerfbusters: Removing ghostly artifacts from casually captured nerfs.
\newblock In \emph{International Conference on Computer Vision (ICCV)}, 2023.

\bibitem[Weber et~al.(2024)Weber, Holynski, Jampani, Saxena, Snavely, Kar, and Kanazawa]{weber2023nerfiller}
Ethan Weber, Aleksander Holynski, Varun Jampani, Saurabh Saxena, Noah Snavely, Abhishek Kar, and Angjoo Kanazawa.
\newblock Nerfiller: Completing scenes via generative 3d inpainting.
\newblock In \emph{Conference on Computer Vision and Pattern Recognition (CVPR)}, 2024.

\bibitem[Wei et~al.(2021)Wei, Liu, Rao, Zhao, Lu, and Zhou]{wei2021nerfingmvs}
Yi Wei, Shaohui Liu, Yongming Rao, Wang Zhao, Jiwen Lu, and Jie Zhou.
\newblock Nerfingmvs: Guided optimization of neural radiance fields for indoor multi-view stereo.
\newblock In \emph{International Conference on Computer Vision (ICCV)}, 2021.

\bibitem[Wu et~al.(2022)Wu, Liu, Chen, Li, Zheng, Cai, and Zheng]{wu2022object}
Qianyi Wu, Xian Liu, Yuedong Chen, Kejie Li, Chuanxia Zheng, Jianfei Cai, and Jianmin Zheng.
\newblock Object-compositional neural implicit surfaces.
\newblock In \emph{European Conference on Computer Vision (ECCV)}, 2022.

\bibitem[Wu et~al.(2023)Wu, Wang, Li, Zheng, and Cai]{wu2023objsdfplus}
Qianyi Wu, Kaisiyuan Wang, Kejie Li, Jianmin Zheng, and Jianfei Cai.
\newblock Objectsdf++: Improved object-compositional neural implicit surfaces.
\newblock In \emph{International Conference on Computer Vision (ICCV)}, 2023.

\bibitem[Wu et~al.(2024)Wu, Mildenhall, Henzler, Park, Gao, Watson, Srinivasan, Verbin, Barron, Poole, and Holynski]{wu2023reconfusion}
Rundi Wu, Ben Mildenhall, Philipp Henzler, Keunhong Park, Ruiqi Gao, Daniel Watson, Pratul~P. Srinivasan, Dor Verbin, Jonathan~T. Barron, Ben Poole, and Aleksander Holynski.
\newblock Reconfusion: 3d reconstruction with diffusion priors.
\newblock In \emph{Conference on Computer Vision and Pattern Recognition (CVPR)}, 2024.

\bibitem[Xiang et~al.(2024)Xiang, Lv, Xu, Deng, Wang, Zhang, Chen, Tong, and Yang]{xiang2024structured}
Jianfeng Xiang, Zelong Lv, Sicheng Xu, Yu Deng, Ruicheng Wang, Bowen Zhang, Dong Chen, Xin Tong, and Jiaolong Yang.
\newblock Structured 3d latents for scalable and versatile 3d generation.
\newblock \emph{arXiv preprint arXiv:2412.01506}, 2024.

\bibitem[Yang et~al.(2023)Yang, Pavone, and Wang]{yang2023freenerf}
Jiawei Yang, Marco Pavone, and Yue Wang.
\newblock Freenerf: Improving few-shot neural rendering with free frequency regularization.
\newblock In \emph{Conference on Computer Vision and Pattern Recognition (CVPR)}, 2023.

\bibitem[Yang et~al.(2024)Yang, Jia, Zhi, and Huang]{yang2024physcene}
Yandan Yang, Baoxiong Jia, Peiyuan Zhi, and Siyuan Huang.
\newblock Physcene: Physically interactable 3d scene synthesis for embodied ai.
\newblock In \emph{Conference on Computer Vision and Pattern Recognition (CVPR)}, 2024.

\bibitem[Yariv et~al.(2021)Yariv, Gu, Kasten, and Lipman]{yariv2021volume}
Lior Yariv, Jiatao Gu, Yoni Kasten, and Yaron Lipman.
\newblock Volume rendering of neural implicit surfaces.
\newblock In \emph{Advances in Neural Information Processing Systems (NeurIPS)}, 2021.

\bibitem[Ye et~al.(2024)Ye, He, Lin, Sheng, Fan, Han, Hu, Yi, Wen, Liu, and Wang]{ye2024pvprecon}
Sheng Ye, Yuze He, Matthieu Lin, Jenny Sheng, Ruoyu Fan, Yiheng Han, Yubin Hu, Ran Yi, Yu-Hui Wen, Yong-Jin Liu, and Wenping Wang.
\newblock Pvp-recon: Progressive view planning via warping consistency for sparse-view surface reconstruction.
\newblock \emph{arXiv preprint arXiv:2409.05474}, 2024.

\bibitem[Yeshwanth et~al.(2023)Yeshwanth, Liu, Nie{\ss}ner, and Dai]{yeshwanthliu2023scannetpp}
Chandan Yeshwanth, Yueh-Cheng Liu, Matthias Nie{\ss}ner, and Angela Dai.
\newblock Scannet++: A high-fidelity dataset of 3d indoor scenes.
\newblock In \emph{International Conference on Computer Vision (ICCV)}, 2023.

\bibitem[Younes et~al.(2024)Younes, Ouasfi, and Boukhayma]{younes2024sparsecraft}
Mae Younes, Amine Ouasfi, and Adnane Boukhayma.
\newblock Sparsecraft: Few-shot neural reconstruction through stereopsis guided geometric linearization.
\newblock In \emph{European Conference on Computer Vision (ECCV)}, 2024.

\bibitem[Young(2021)]{xatlas}
Jonathan Young.
\newblock xatlas.
\newblock \url{https://github.com/jpcy/xatlas}, 2021.

\bibitem[Yu et~al.(2023)Yu, Guo, Li, Liang, Zhang, and Qi]{yu2023text}
Xin Yu, Yuan-Chen Guo, Yangguang Li, Ding Liang, Song-Hai Zhang, and Xiaojuan Qi.
\newblock Text-to-3d with classifier score distillation.
\newblock \emph{arXiv preprint arXiv:2310.19415}, 2023.

\bibitem[Yu et~al.(2022)Yu, Peng, Niemeyer, Sattler, and Geiger]{yu2022monosdf}
Zehao Yu, Songyou Peng, Michael Niemeyer, Torsten Sattler, and Andreas Geiger.
\newblock Monosdf: Exploring monocular geometric cues for neural implicit surface reconstruction.
\newblock In \emph{Advances in Neural Information Processing Systems (NeurIPS)}, 2022.

\bibitem[Zhang et~al.(2021)Zhang, Cui, Zhang, Zeng, Pollefeys, and Liu]{Zhang_2021_im3d}
Cheng Zhang, Zhaopeng Cui, Yinda Zhang, Bing Zeng, Marc Pollefeys, and Shuaicheng Liu.
\newblock Holistic 3d scene understanding from a single image with implicit representation.
\newblock In \emph{Conference on Computer Vision and Pattern Recognition (CVPR)}, 2021.

\bibitem[Zhang et~al.(2020)Zhang, Riegler, Snavely, and Koltun]{zhang2020nerf++}
Kai Zhang, Gernot Riegler, Noah Snavely, and Vladlen Koltun.
\newblock Nerf++: Analyzing and improving neural radiance fields.
\newblock \emph{arXiv preprint arXiv:2010.07492}, 2020.

\bibitem[Zhang et~al.(2023)Zhang, Rao, and Agrawala]{zhang2023adding}
Lvmin Zhang, Anyi Rao, and Maneesh Agrawala.
\newblock Adding conditional control to text-to-image diffusion models.
\newblock In \emph{International Conference on Computer Vision (ICCV)}, 2023.

\bibitem[Zhao et~al.(2025)Zhao, Li, Li, Qi, Ruan, Zhu, and Althoefer]{zhao2025tac}
Zihang Zhao, Yuyang Li, Wanlin Li, Zhenghao Qi, Lecheng Ruan, Yixin Zhu, and Kaspar Althoefer.
\newblock Tac-{M}an: Tactile-informed prior-free manipulation of articulated objects.
\newblock \emph{Transactions on Robotics (T-RO)}, 2025.

\bibitem[Zhou et~al.(2024)Zhou, Ran, Xiong, He, Lin, Wang, Sun, and Yang]{zhou2024gala3d}
Xiaoyu Zhou, Xingjian Ran, Yajiao Xiong, Jinlin He, Zhiwei Lin, Yongtao Wang, Deqing Sun, and Ming-Hsuan Yang.
\newblock Gala3d: Towards text-to-3d complex scene generation via layout-guided generative gaussian splatting.
\newblock In \emph{International Conference on Machine Learning (ICML)}, 2024.

\bibitem[Zhou and Tulsiani(2023)]{zhou2023sparsefusion}
Zhizhuo Zhou and Shubham Tulsiani.
\newblock Sparsefusion: Distilling view-conditioned diffusion for 3d reconstruction.
\newblock In \emph{Conference on Computer Vision and Pattern Recognition (CVPR)}, 2023.

\end{thebibliography}

\clearpage
\appendix
\renewcommand{\thefigure}{S.\arabic{figure}}
\renewcommand{\thetable}{S.\arabic{table}}
\renewcommand{\theequation}{S.\arabic{equation}}
\maketitlesupplementary

\noindent We provide details on the decompositional reconstruction process, training procedures, experimental setup, and a discussion of limitations. Additionally, we highly recommend watching the demo video on our webpage for a more intuitive and visually engaging presentation of the results.

\section{Generalizability to in-the-wild videos}
\begin{figure*}[htbp]
    \centering
    \includegraphics[width=\linewidth]{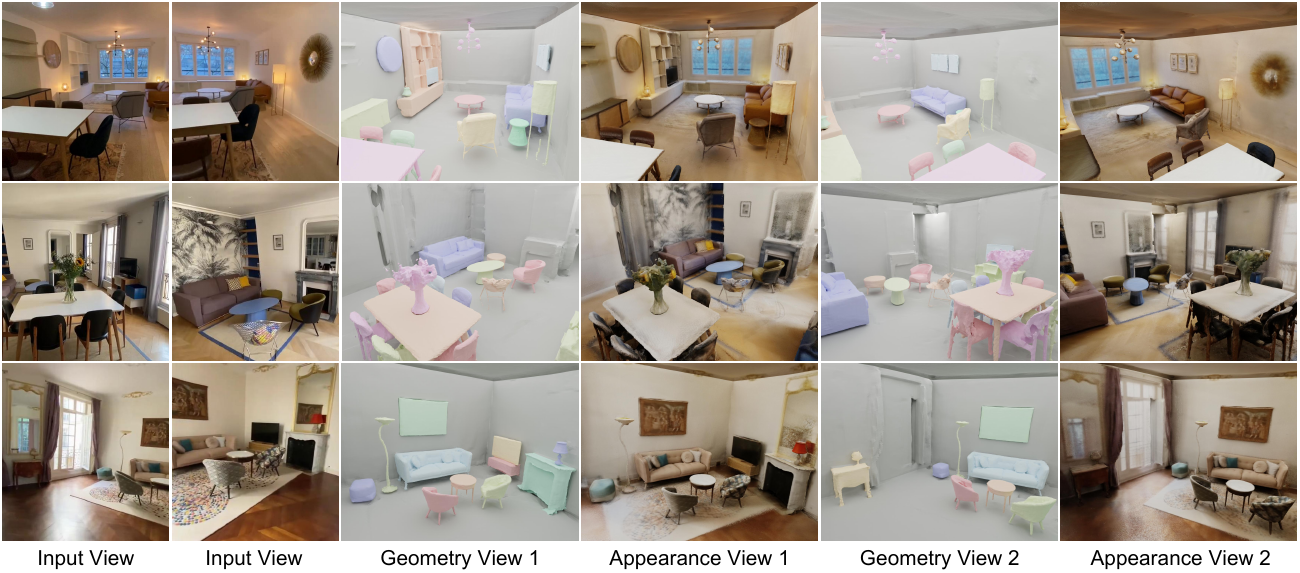}
    \caption{\textbf{Generalizability to YouTube videos with 15 input views.} The reconstruction results highlight our model's robust ability to generalize to in-the-wild indoor scenes.}
    \vspace{-0.1in}
    \label{fig:youtube_demo}
\end{figure*}

Our method demonstrates robust generalizability to in-the-wild indoor scenes. \cref{fig:youtube_demo} presents reconstruction results of in-the-wild YouTube videos using 15 input views, with camera poses calibrated via COLMAP~\cite{schoenberger2016sfm} and object masks obtained from SAM2~\cite{ravi2024sam2}.

\section{Decompositional Reconstruction}

We first present preliminary of decompositional scene reconstruction, also known as object-compositional reconstruction, which aims to reconstruct each object in the scene individually—including both foreground objects and the background—rather than representing the entire scene as a single, inseparable mesh.

\subsection{Decompositional Representation}
Following previous work~\cite{wu2022object,wu2023objsdfplus,li2023rico,ni2024phyrecon}, we utilize a set of posed RGB images and their corresponding instance masks to achieve decompositional reconstruction of objects, treating the background also as an object.

As described in the main paper, for a scene with \(k\) objects, we predict \(k\) \ac{sdf}s for each point \(\bm{p}\) and the \(j\)-th \( (1 \leq j \leq k) \) SDF \(s_j(\bm{p})\) is for the \(j\)-th object. The scene SDF \(s(\bm{p})\) is the minimum of the object \ac{sdf}s:

\begin{equation}
\label{eq:scene_sdf}
    s(\bm{p}) = \min_{1\leq j\leq k}~s_j(\bm{p}), 
\end{equation}

The normal \(\bm{n}(\bm{p})\) is the gradient of \(s(\bm{p})\), while normal \(\bm{n}^j(\bm{p})\) is the gradient of \(s_j(\bm{p})\):
\begin{equation}
\label{eq:normal_supp}
    \bm{n}(\bm{p}) = \nabla s(\bm{p}),
    \quad \bm{n}^j(\bm{p}) = \nabla s_j(\bm{p})
\end{equation}

Next, we transform each point's \acs{sdf}s into instance semantic logits $\bm{h}(\bm{p}) = [h_1(\bm{p}), h_2(\bm{p}), \ldots, h_k(\bm{p})]$, where
\begin{equation}
\label{eq:semantic_logits}
\begin{aligned}
h_j(\bm{p}) &= \gamma / (1 + \exp(\gamma \cdot s_j(\bm{p}))), 
\end{aligned}
\end{equation}
where \(\gamma=10\) is a fixed parameter in our implementation.

\subsection{Volume Rendering}

For each camera ray \(\bm{r}=(\bm{o},\bm{d})\) with \(\bm{o}\) as the ray origin (camera center) and \(\bm{d}\) as the viewing direction, \(n\) points \(\left\{\bm{p}_{i}=\bm{o}+t_{i}\bm{d} \mid i=0,1,\ldots,n-1\right\}\) are sampled, where \(t_{i}\) is the distance from the sample point to the camera center. We predict \(k\) \ac{sdf}s and the color \(\bm{c}_{i}\) for each point \(\bm{p}_{i}\) along the ray. Then we compute scene \ac{sdf} \(s_i\), normal \(\bm{n}_i\) and instance sematic logits \(\bm{h}_i\) for point \(\bm{p}_{i}\) by \cref{eq:scene_sdf,eq:normal_supp,eq:semantic_logits}.
Next, we convert the scene \ac{sdf} \(s(\bm{p})\) into density \(\sigma\) for volume rendering as in NeRF~\cite{mildenhall2020nerf} following the method introduced in VolSDF~\cite{yariv2021volume}:
\begin{equation}
\label{eq:sdf_to_density_supp}
    \sigma(s) = \begin{cases} \frac{1}{2\beta}\exp(\frac{s}{\beta}) & s \leq 0 \\ \frac{1}{\beta}(1 - \exp(-\frac{s}{\beta})) & s \ge 0 \end{cases},
\end{equation}
where \(\beta\) is a learnable parameter.
We then calculate the discrete accumulated transmittance \(T_i\) and discrete opacity \(\alpha_i\) as follows:
\begin{equation}
\label{eq:transmittance_supp}
    T_{i} = \prod_{j=0}^{i-1}(1-\alpha_{j}),
    \quad \alpha_i = 1-exp(-\sigma_i\delta_i),
\end{equation}
where \(\delta_i\) represents the distance between neighboring sample points along the ray.

Using volume rendering, the predicted scene color \(\hat{\bm{C}}(\bm{r})\), depth \(\hat{D}(\bm{r})\), normal \(\hat{\bm{N}}(\bm{r})\) and instance semantic \(\hat{\bm{S}}(\bm{r})\) for the ray \(r\) are computed as:

\begin{equation}
\label{eq:volume_rendering_color_supp}
\begin{split}
    \hat{\bm{C}}(\bm{r}) & = \sum_{i=0}^{n-1} T_{i}\alpha_{i}\bm{c}_{i}, \quad \hat{D}(\bm{r}) = \sum_{i=0}^{n-1} T_{i}\alpha_{i}t_{i}, \\
    \hat{\bm{N}}(\bm{r}) & = \sum_{i=0}^{n-1} T_{i}\alpha_{i}\bm{n}_{i}, \quad \hat{\bm{S}}(\bm{r}) = \sum_{i=0}^{n-1} T_{i}\alpha_{i}\bm{h}_{i},
\end{split}
\end{equation}

Additionally, replacing the scene \ac{sdf} \(s\) with \(j\)-th object \ac{sdf} \(s_j\) in \cref{eq:sdf_to_density_supp,eq:transmittance_supp} allows rendering of the normal \(\hat{\bm{N}}^j(\bm{r})\) and mask \(\hat{O}^j(\bm{r})\) for \(j\)-th object as:
\begin{equation}
\label{eq:object_volume_rendering}
    \hat{\bm{N}}^j(\bm{r}) = \sum_{i=0}^{n-1} T^j_{i}\alpha^j_{i}\bm{n}^j_{i}, \quad \hat{O}^j(\bm{r}) = \sum_{i=0}^{n-1} T^j_{i}\alpha^j_{i},
\end{equation}

\subsection{Loss function}
\label{sec:loss_recon}

\paragraph{RGB Reconstruction Loss}
Given input images, we employ RGB reconstruction loss \(\mathcal{L}_{C}\) to minimize the difference between ground-truth pixel color and the rendered color. We follow the Yu \etal~\cite{yu2022monosdf} here for the RGB reconstruction loss:

\begin{equation}
\label{eq:color_rendering_loss}
    \mathcal{L}_{C} = \sum_{\bm{r} \in \mathcal{R}}||\hat{\bm{C}}(\bm{r}) - \bm{C}(\bm{r})||_1,
\end{equation}
where $\hat{\bm{C}}(\bm{r})$ is the rendered color from volume rendering and $\bm{C}(\bm{r})$ denotes the ground truth.

\paragraph{Depth Consistency Loss}
Monocular depth and normal cues~\cite{yu2022monosdf} can greatly benefit indoor scene reconstruction. For the depth consistency, we minimize the difference between rendered depth $\hat{D}(\bm{r})$ and the depth estimation $\bar{D}(\bm{r})$ from the Depthfm model~\cite{gui2024depthfm}:

\begin{equation}
    \mathcal{L}_{D} =  \sum_{\bm{r} \in \mathcal{R}}|| (w\hat{D}(\bm{r}) + q) - \bar{D}(\bm{r}) ||^2,
\end{equation}
where $w$ and $q$ are the scale and shift values to match the different scales. We solve $w$ and $q$ with a least-squares criterion, which has the closed-form solution. Please refer to the supplementary material of~\cite{yu2022monosdf} for a detailed computation process.

\paragraph{Normal Consistency Loss}
We utilize the normal cues $\bar{\bm{N}}(\bm{r})$ from Omnidata model~\cite{eftekhar2021omnidata} to supervise the rendered normal through the normal consistency loss, which comprises L1 and angular losses:
\begin{equation}
    \mathcal{L}_{N} = \sum_{\bm{r} \in \mathcal{R}} ||\hat{\bm{N}}(\bm{r}) - \bm{\bar{N}}(\bm{r})||_1 + ||1- \hat{\bm{N}}(\bm{r})^{T}\bm{\bar{N}}(\bm{r})||_1.
\end{equation}
The rendered normal \(\hat{\bm{N}}(\bm{r})\) and normal cues \(\bm{\bar{N}}(\bm{r})\) will be transformed into the same coordinate system by the camera pose.

\paragraph{Instance Semantic Loss}
We minimize the semantic loss between rendered instance semantic logits of each pixel and the ground-truth pixel instance class. The instance semantic objective is implemented as a cross-entropy loss:
\begin{equation}
    \mathcal{L}_{S} = \sum_{\bm{r} \in \mathcal{R}}\sum_{j=1}^k-\bar{h}_j(\bm{r})\log h_j(\bm{r}).
\end{equation}
The $\bar{h}_j(\bm{r})$ is the ground-truth semantic probability for $j$-th object, which is $1$ or $0$.

\paragraph{Eikonal Loss and Smoothness Loss}
Following common practice~\cite{yariv2021volume,oechsle2021unisurf}, we add an Eikonal regularization and smoothness regularization term on the sampled points to regularize the SDF learning by:

\begin{equation}
    \mathcal{L}_{eik} = \sum_{i=0}^{n-1}(|| \nabla s(\bm{p}_i)||_2 - 1),
\end{equation}

\begin{equation}
    \mathcal{L}_{smo} = \sum_{i=0}^{n-1}( || \nabla s(\bm{p}_i) - \nabla s(\bm{\tilde{p}}_i)||_1),
\end{equation}
where \(\bm{\tilde{p}}_i\) is randomly sampled nearby the \(\bm{p}_i\).

\paragraph{Background Smoothness Loss}
Following the previous work RICO~\cite{li2023rico}, we use background smoothness loss to regularize the geometry of the occluded background to be smooth. Specifically, we randomly sample a $\mathcal{P}\times \mathcal{P}$ size patch every $\mathcal{T}_{\mathcal{P}}$ iterations within the given image and compute semantic map $\hat{\bm{H}}(\bm{r})$ and a patch mask $\hat{M}(\bm{r})$:
\begin{equation}
    \hat{M}(\bm{r}) = \mathds{1}[\arg\max (\hat{\bm{H}}(\bm{r})) \neq 1],
\end{equation}
wherein the mask value is $1$ if the rendered class is not the background, thereby ensuring only the occluded background is regulated. Subsequently, we calculate the background depth map $\bar{D}(\bm{r})$ and background normal map $\bar{\bm{N}}(\bm{r})$ using the background SDF exclusively. The patch-based background smoothness loss is then computed as:
\begin{equation}
\begin{split}
    \mathcal{L}(\hat{D}) & = \sum_{d=0} ^3 \sum_{m,n=0}^{\mathcal{P}-1-2^d} \hat{M}(\bm{r}_{m,n}) \odot (|\hat{D}(\bm{r}_{m,n}) - \\
    & \hat{D}(\bm{r}_{m,n+2^d})| + |\hat{D}(\bm{r}_{m,n}) - \hat{D}(\bm{r}_{m+2^d,n})| ),
\end{split}
\end{equation}

\begin{equation}
\begin{split}
    \mathcal{L}(\hat{\bm{N}}) & = \sum_{d=0} ^3 \sum_{m,n=0}^{\mathcal{P}-1-2^d} \hat{M}(\bm{r}_{m,n}) \odot (|\hat{\bm{N}}(\bm{r}_{m,n}) - \\
    & \hat{\bm{N}}(\bm{r}_{m,n+2^d})|  + |\hat{\bm{N}}(\bm{r}_{m,n}) - \hat{\bm{N}}(\bm{r}_{m+2^d,n})| ),  
\end{split}
\end{equation}

\begin{equation}
\mathcal{L}_{bs} = \mathcal{L}(\hat{D}) + \mathcal{L}(\hat{\bm{N}})
\end{equation}

\paragraph{Object Distinction Regularization Loss}
Following \objectsdfpp~\cite{wu2023objsdfplus}, we employ a regularization term on object \ac{sdf}s to penalize the overlap between any two objects:
\begin{equation}
    \mathcal{L}_{sdf} = \sum_{i=0}^{n-1}( \sum_{j=1}^{k} ReLU(-s_j(\bm{p}_i) - s(\bm{p}_i))).
\end{equation}

\begin{figure*}[ht!]
    \centering
    \includegraphics[width=\linewidth]{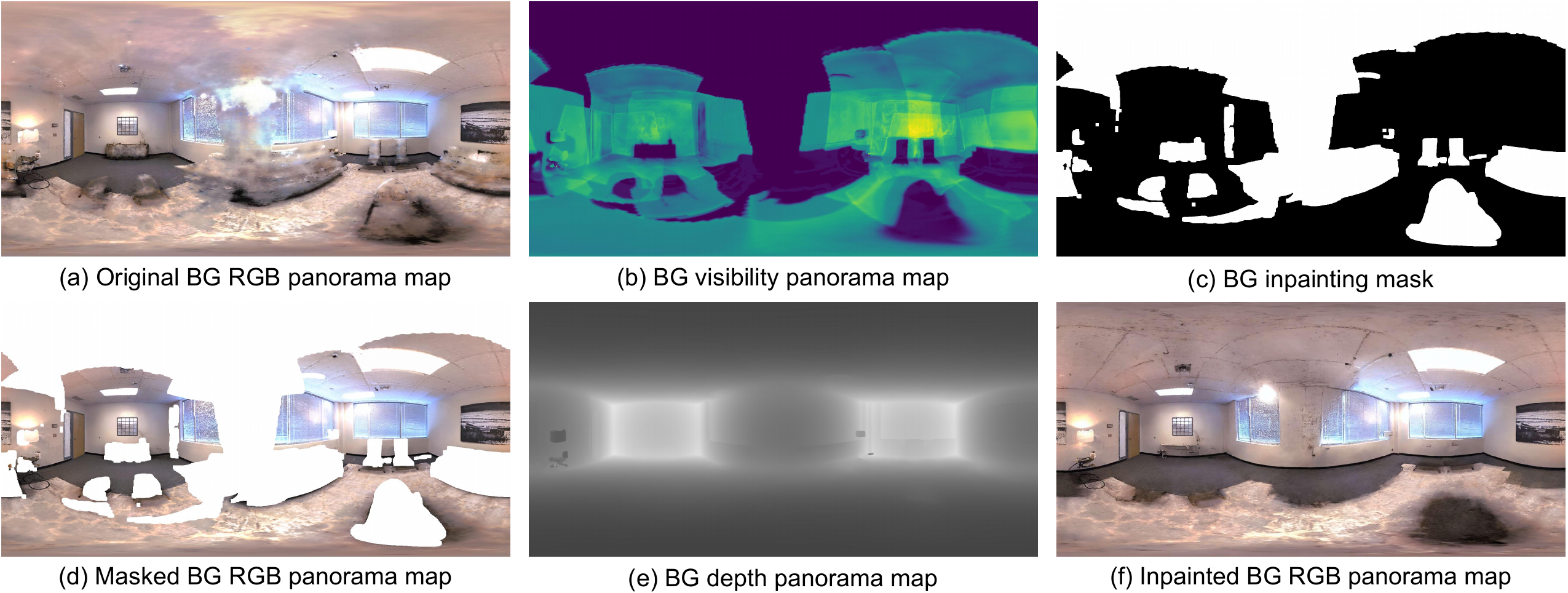}
    \caption{\textbf{Panorama inpainting of background}. We present the original panorama map, the inpainted panorama map, the visibility panorama map, as well as the mask and depth guidance used in the depth-guided panorama inpainting process.}
    \label{fig:bg_inpaint}
\end{figure*}

\section{More Training Details}

\subsection{Object-compositional Reconstruction}
\paragraph{Overall Loss}
We employ all the aforementioned losses in \cref{sec:loss_recon} during the object-compositional reconstruction stage:

\begin{equation}
\label{eq:total_recon_loss}
    \begin{split}
        \mathcal{L}_{recon} & = \mathcal{L}_{C} + \lambda_D \mathcal{L}_{D} + \lambda_N\mathcal{L}_{N} + \lambda_S\mathcal{L}_{S} + \lambda_{bs}\mathcal{L}_{bs}  \\
        & + \lambda_{eik}\mathcal{L}_{eik} + \lambda_{smo}\mathcal{L}_{smo} + \lambda_{sdf}\mathcal{L}_{sdf},
    \end{split}
\end{equation}
where the loss weights are set as \(\lambda_D=0.1\), \(\lambda_N=0.05\), \(\lambda_S=1.0\), \(\lambda_{bs}=0.1\), \(\lambda_{eik}=0.1\), \(\lambda_{smo}=0.005\), \(\lambda_{sdf}=0.5\) following previous work~\cite{yu2022monosdf,li2023rico,wu2023objsdfplus}.

\paragraph{Optimization of Visibility Grid}
At the end of the object-compositional reconstruction stage, when the transmittance achieves sufficient accuracy, we optimize the visibility grid \(G\). During this process, all input views are rendered \(M\) times using ~\cref{eq:volume_rendering_color_supp}, and the accumulated transmittance \(T_i\) is utilized to optimize the visibility value of point \(\bm{p}_i\). Note that \(M\) is a hyperparameter that impacts the final visibility grid values; in our implementation, we set \(M=20\). After the optimization, the visibility grid is frozen for the subsequent geometry and appearance optimization stages.

\subsection{Geometry Optimization}
\paragraph{Input for the Diffusion Model}
In the common case, the encoder of Stable Diffusion~\cite{rombach2022high} is used to encode an image into the latent code \(z\) during \ac{sds}, followed by employing the UNet of Stable Diffusion to predict the score \(\hat{\epsilon}\) and compute the \ac{sds} loss. However, the encoding process is relatively slow in computational speed. To address this issue and facilitate efficient training, following the previous work~\cite{chen2023fantasia3d,qiu2024richdreamer}, at each training iteration, we directly downsample the concatenated map \(\tilde{n}_j\), which consists of the normal map \(\hat{\bm{N}}^j(\bm{r})\) and mask map \(\hat{O}^j(\bm{r})\) rendered by \cref{eq:object_volume_rendering} for sampled \(j\)-th object, into the latent code \(z\) dimension. This approach reduces the computation time by approximately half without causing any performance degradation compared to directly inputting the normal map \(\hat{\bm{N}}^j(\bm{r})\) into the encoder.

\paragraph{Overall Loss}
We employ reconstruction loss \(\mathcal{L}_{recon}\) in \cref{eq:total_recon_loss} and visibility-guided geometry \ac{sds} loss \(\mathcal{L}_{\text{SDS}}^{g-v}\) in the geometry optimization stage:

\begin{equation}
    \mathcal{L}_{geo} = \mathcal{L}_{recon} + \lambda_{sds}^{geo} \mathcal{L}_{\text{SDS}}^{g-v},
\end{equation}
where \(\lambda_{sds}^{geo}=1e^{-5}\) in our implementation.

\begin{figure}[ht!]
    \centering
    \includegraphics[width=\linewidth]{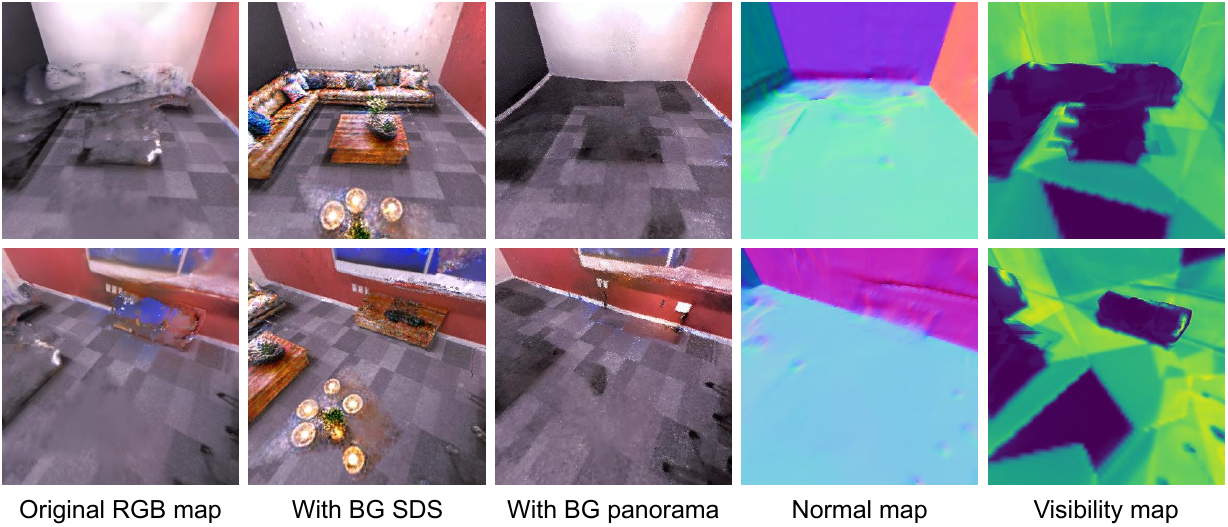}
    \caption{\textbf{Failure case of directly optimizing background with \ac{sds} loss.} Incorporating appearance \ac{sds} for the background may result in hallucinated non-existent objects in low-visibility regions, even when the smooth normal map indicates no objects in the background. On the contrary, the background map obtained after optimizing depth-guided panorama inpainting produces a cleaner and more reasonable background texture.}
    \label{fig:bg_sds_failure_case}
\end{figure}

\subsection{Appearance Optimization}

\paragraph{Rendering Loss in Appearance Optimization Stage}
We employ two types of color rendering losses during the appearance optimization stage to ensure consistency between the observed regions and the input views. The first type derives directly from the input views, where we apply \(\mathcal{L}_C\) as defined in \cref{eq:color_rendering_loss}.
The second type leverages useful appearance information distilled from the reconstruction network. For this, we randomly sample camera views within the scene, render RGB and visibility maps, and use regions with high visibility for appearance supervision.
The sum of these two color rendering losses is denoted as \(\mathcal{L}_C^{a}\).

\paragraph{Depth-guided Panorama Inpainting and Loss for BG}
Applying appearance \ac{sds} \(\mathcal{L}_{\text{SDS}}^{a-v}\) for background appearance optimization often leads to degenerated results, \eg introducing non-existent objects as shown in \cref{fig:bg_sds_failure_case}, due to the lack of clear geometric cues in the background from the local camera perspective. To address this issue, inspired by previous work~\cite{wang2024roomtex,matterport2024defurnishing}, we adopt depth-guided inpainting~\cite{zhang2023adding} to refine the low-visibility regions of the background panorama color map. 
Specifically, we first generate the original RGB, visibility, and depth panorama maps, as shown in \cref{fig:bg_inpaint} (a, b, e). Next, we obtain the inpainting mask (\cref{fig:bg_inpaint} (c)) for regions where the visibility map falls below a threshold \(\tau\) (set to \(\tau=0.2\) in our implementation). Finally, we apply depth-guided inpainting to produce the inpainted RGB panorama map (\cref{fig:bg_inpaint} (f)).

To supervise the background appearance during the appearance optimization stage, we transform the inpainted RGB panorama map into a set of perspective images with corresponding camera poses. At each training iteration, we sample \(B\) perspective images \(\bm{C}_B\) along with their associated camera poses. For these poses, we render the background color maps \(\hat{\bm{C}}_B\), and define the background panorama loss as:

\begin{equation}
\label{eq:bg_pano_loss}
    \mathcal{L}_{\text{bg-pano}} = \frac{1}{B} \sum_{i=1}^{B}||\hat{\bm{C}}_B - \bm{C}_B||_1,
\end{equation}

\paragraph{Overall Loss}
We employ color rendering loss \(\mathcal{L}_C^{a}\), background panorama loss \(\mathcal{L}_{\text{bg-pano}}\) in \cref{eq:bg_pano_loss} and visibility-guided appearance \ac{sds} loss \(\mathcal{L}_{\text{SDS}}^{a-v}\) in the appearance optimization stage:

\begin{equation}
    \mathcal{L}_{app} = \lambda_C^a \mathcal{L}_C^{a} + \lambda_{\text{bg-pano}} \mathcal{L}_{\text{bg-pano}} + \mathcal{L}_{\text{SDS}}^{a-v},
\end{equation}
where \(\lambda_C^a=\lambda_{\text{bg-pano}}=1e^{4}\) in our implementation.

\begin{figure}[ht!]
    \centering
    \includegraphics[width=\linewidth]{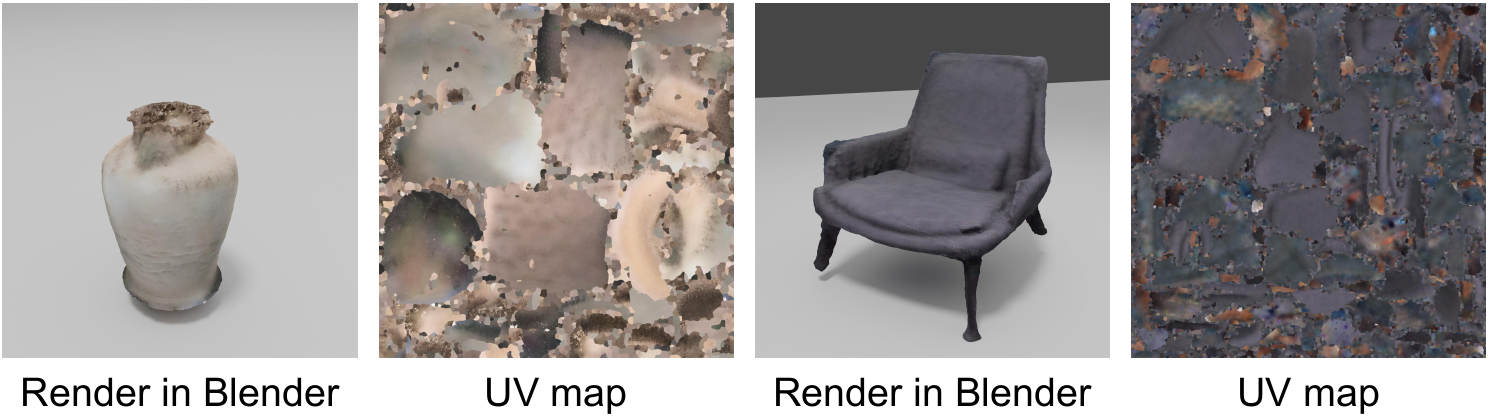}
    \caption{\textbf{Visualization of UV mapping and rendering results.} Our method produces completed meshes with detailed UV maps, enabling photorealistic rendering in common 3D software such as Blender.}
    \label{fig:uv_map}
\end{figure}

\paragraph{Export UV Map}
Following previous works~\cite{chen2023fantasia3d,qiu2024richdreamer}, we utilize the trained \(\psi\) to export the appearance of object mesh as UV map by the xatlas~\cite{xatlas}. Our object mesh with exported UV map supports direct use and editing in common 3D software, \eg Blender~\cite{blender}, as shown in \cref{fig:uv_map}. We export \(1024 \times 1024\) UV map for foreground objects and \(2048 \times 2048\) UV map for the background in our case.

\section{More Experiment Details}

\subsection{Baselines Details}
For MonoSDF~\cite{yu2022monosdf}, \rico~\cite{li2023rico}, and \objectsdfpp~\cite{wu2023objsdfplus}, which are designed for reconstruction, we directly utilize the official code to obtain the reconstructed meshes and rendered images. 
For FreeNeRF~\cite{yang2023freenerf}, which focuses on novel view synthesis and does not include reconstruction code, we first predict depth maps and RGB images for densely sampled camera views within the scene. We then apply TSDF Fusion~\cite{curless1996tsdf} to integrate the predicted depth maps into a TSDF volume and export the resulting mesh.
ZeroNVS~\cite{Kyle2024zeronvs} trains a diffusion model to synthesize novel views of scenes from a single image. To adapt it for multi-view inputs, we follow the approach of ReconFusion~\cite{wu2023reconfusion}, using the input view closest to the novel view as the conditioning view for the diffusion model. We denote the adapted version as ZeroNVS*. Subsequently, we use MonoSDF to reconstruct the scene mesh from the images synthesized by ZeroNVS*.

\subsection{Data Preparation}

\paragraph{Monocular Cues}
We utilize the pre-trained DepthFM model~\cite{gui2024depthfm} and Omnidata model~\cite{eftekhar2021omnidata} to generate the depth map $\bar{D}$ and normal map $\bar{\bm{N}}$ for each RGB image, respectively. 
While depth cues provide semi-local relative information, normal cues are inherently local, capturing fine geometric details. As a result, we expect surface normals and depth to complement each other effectively. 
It is worth noting that estimating absolute scale in general scenes remains challenging; therefore, $\bar{D}$ should be interpreted as a relative depth cue.

\paragraph{GT Instance Mask}
For the ScanNet++~\cite{yeshwanthliu2023scannetpp} dataset, we use the official rendering engine to generate instance masks for each image based on the provided GT mesh and per-vertex 3D instance annotations. For the Relica~\cite{replica19arxiv} datasets,  we utilize the original instance masks from vMAP~\cite{kong2023vmap}, which are overly fragmented, and manually merge adjacent fragmented instance masks into coherent objects. 
Furthermore, since both ScanNet++ and Replica only provide a complete mesh of the scene, we derive the background GT mesh by removing the object meshes from the total mesh and manually filling the holes.

Notably, with the rapid advancement of segmentation and tracking models, such as SAM~\cite{kirillov2023segany} and SAM2~\cite{ravi2024sam2}, it's more feasible to extract object masks directly from images using off-the-shelf models. These tools could inspire further progress in decompositional neural scene reconstruction.

\subsection{Reconstruction Metrics Details}
Following previous research~\cite{yu2022monosdf,li2023rico,wu2023objsdfplus}, we evaluate the \ac{cd} in $cm$, \acs{fscore} with a threshold of $5cm$ and \ac{nc} for 3D scene and object reconstruction. Consistent with previous studies~\cite{yu2022monosdf,li2023rico,wu2023objsdfplus}, reconstruction is evaluated only on visible areas for the entire scene, while complete meshes are assessed for individual objects and background meshes.
These metrics are defined in \cref{tab:metrics_definition}.

Since the baselines ZeroNVS*~\cite{Kyle2024zeronvs}, FreeNeRF~\cite{yang2023freenerf} and MonoSDF~\cite{yu2022monosdf} can only reconstruct the total scene and cannot decompose it into individual objects, we evaluate the metrics only for the total scene, \ie, the total scene reconstruction metrics and rendering metrics.

\begin{table*}[ht!]
\caption{\textbf{Evaluation metrics.} We show the evaluation metrics with their definitions that we use to measure reconstruction quality. $P$ and $P^*$ are the point clouds sampled from the predicted and the ground truth mesh. $\bm{n}_{\bm{p}}$ is the normal vector at point $\bm{p}$.}
\small
\centering
    \setlength\extrarowheight{10pt}
    \begin{tabular}{lc}
    \toprule
    Metric & Definition \\
    \midrule
    \textbf{\acf{cd}}  & $\frac{\textit{Accuracy} + \text{Completeness}}{2} $ \\
    \textit{Accuracy} & $\underset{\bm{p} \in P}{\mbox{mean}}\left( \underset{\bm{p}^*\in P^*}{\mbox{min}} ||\bm{p}-\bm{p}^*||_1 \right)$ \\
    \textit{Completeness} & $\underset{\bm{p}^* \in P^*}{\mbox{mean}}\left( \underset{\bm{p} \in P}{\mbox{min}} ||\bm{p}-\bm{p}^*||_1 \right)$\\
    \midrule
    \textbf{F-score} & $\frac{ 2 \times \text{Precision} \times \text{Recall} }{\text{Precision} + \text{Recall}}$ \\
    \textit{Precision} & $\underset{\bm{p} \in P}{\mbox{mean}}\left( \underset{\bm{p}^*\in P^*}{\mbox{min}} ||\bm{p}-\bm{p}^*||_1 < 0.05 \right)$  \\
    \textit{Recall} & $\underset{\bm{p}^* \in P^*}{\mbox{mean}}\left( \underset{\bm{p} \in P}{\mbox{min}} ||\bm{p}-\bm{p}^*||_1 < 0.05\right)$ \\
    \midrule
    \textbf{Normal Consistency} & $ \frac{\textit{Normal Accuracy} + \textit{Normal Completeness}}{2} $\\
    \textit{Normal Accuracy}  &  $\underset{\bm{p} \in P}{\mbox{mean}}\left(  \bm{n}_{\bm{p}}^T\bm{n}_{\bm{p}^*}  \right) \,\, \text{s.t.} \, \, \bm{p}^* = \underset{\bm{p}^* \in P^*}{arg\,min} ||\bm{p}-\bm{p}^*||_1 $  \\
    \textit{Normal Completeness} &  $\underset{\bm{p}^* \in P^*}{\mbox{mean}}\left( \bm{n}_{\bm{p}}^T \bm{n}_{\bm{p}^*} \right) \,\, \text{s.t.} \, \, \bm{p} = \underset{\bm{p} \in P}{arg\,min} ||\bm{p}-\bm{p}^*||_1 $  \\
    \bottomrule
    \end{tabular}
    \vspace{.1cm}
\label{tab:metrics_definition}
\end{table*}

\begin{table*}[h!]
\caption{\textbf{Training time and performance comparison.} Our method outperforms baselines in 4.5 hours per scene with 50,000 iterations.}
\label{tab:running_time}
\centering
\resizebox{0.75\linewidth}{!}{
\begin{tabular}{cccccccccc}
\toprule
\multirow{2}{*}{\textbf{Total Iter}} & \multicolumn{3}{c}{\textbf{RICO}} & \multicolumn{3}{c}{\textbf{ObjectSDF++}} & \multicolumn{3}{c}{\textbf{Ours}} \\
\cmidrule(lr){2-4} \cmidrule(lr){5-7} \cmidrule(lr){8-10}
 & Time & CD$\downarrow$ & F-Score$\uparrow$ & Time & CD$\downarrow$ & F-Score$\uparrow$ & Time & CD$\downarrow$ & F-Score$\uparrow$ \\
\midrule
40000 & 2.63h & 21.30 & 49.47 & 2.34h & 18.48 & 52.51 & 2.59h & 12.96 & 61.87 \\
50000 & 3.16h & 17.37 & 52.33 & 2.93h & 13.36 & 60.37 & \cellcolor{gbest}4.52h & \cellcolor{gbest}4.51 & \cellcolor{gbest}72.66 \\
60000 & 3.82h & 14.63 & 57.74 & 3.42h & 5.42 & 70.19 & 6.54h & 4.35 & 73.23 \\
80000 & \cellcolor{gthird}5.01h & \cellcolor{gthird}12.09 & \cellcolor{gthird}63.39 & \cellcolor{gsecond}4.48h & \cellcolor{gsecond}5.10 & \cellcolor{gsecond}70.87 & 10.45h & 4.33 & 73.32 \\
\bottomrule
\end{tabular}
}
\end{table*}

\subsection{Implementation Details}
We implement our model in PyTorch~\cite{paszke2019pytorch} and utilize the Adam optimizer~\cite{kingma2014adam} with an initial learning rate of $5e-4$. In the object-compositional reconstruction stage, we sample 1024 rays per iteration, and in the geometry and appearance optimization stages, we render \(128 \times 128\) images for normal, mask, and color maps. We use \(2048 \times 1024\) for the background panorama map. 

For visibility guided \ac{sds}:

\begin{equation}
    w^{v}(z) = 
    \begin{cases} 
     w_0 + m_0 V(z) & \text{if } V(z) \leq \tau \\ 
     w_1 + m_1 V(z) & \text{if } V(z) > \tau
    \end{cases},
\end{equation}
we set \(\tau=0.5,w_0=20,m_0=-38,w_1=2,m_1=-2\) for the geometry optimization stage. Under this configuration, \(w^{v}(z)\) achieves a maximum value of \(20\) when \(V(z)=0\), a minimum value of \(0\) when \(V(z)=1\), and a value of \(1\) when \(V(z)=\tau=0.5\).
For appearance optimization stage, we set \(\tau=0.3,w_0=1,m_0=0,w_1=0,m_1=0\), which results in \(w^{v}(z)=1\) when \(V(z) \leq 0.3 \) and \(w^{v}(z)=0\) when \(V(z) > 0.3 \).

Our model is trained for 80000 iterations on both Replica~\cite{replica19arxiv} and ScanNet++~\cite{yeshwanthliu2023scannetpp} datasets. The geometry optimization stage and appearance optimization stage begin at the \(35000^{th}\) and \(75000^{th}\) iterations, respectively.
All experiments are conducted on a single NVIDIA-A100 GPU, requiring approximately 10 hours to complete the training of a single scene.

\subsection{Training Time Comparison}
For a fair comparison with prior work~\cite{yu2022monosdf,wu2023objsdfplus,li2023rico},
we train our model and all baselines for 80,000 iterations in all main paper experiments. Detailed training time and performance results are provided in \cref{tab:running_time}, showing that our method outperforms baselines in approximately 4.5 hours per scene with 50,000 iterations.

\section{Scene Editing Details}

\subsection{Text-based Editing}
With our decompositional representation, which breaks down each object's representation in the scene, we can seamlessly edit the representation of any object in the scene based on the text prompt, with the generative capabilities of our \ac{sds} diffusion prior.
With the two forms of \ac{sds} prior we introduced, \ie, the geometry \ac{sds} prior and the appearance \ac{sds} prior, we can freely edit both the geometry and appearance of objects. During the editing process, we exclude the reconstruction loss for the edited object and disable the visibility guidance.
\paragraph{Geometry Editing}
We realize geometry editing for objects in the geometry optimization stage using geometry \ac{sds} loss \(\mathcal{L}_{\text{SDS}}^{g}\). 
During optimization, we replace the original object prompt with the desired object prompt while continuing to sample novel camera views around the original object's bounding box. This ensures that the desired object retains the same location as the original one. 

\paragraph{Appearance Editing}
We perform object appearance editing during the appearance optimization stage using the appearance \ac{sds} loss \(\mathcal{L}_{\text{SDS}}^{a}\). 
For this task, we not only modify the object prompt but also update the negative prompt in Stable Diffusion~\cite{rombach2022high}, as suggested in the prompt engineering tutorial~\cite{sdxlstyle}. Empirically, we observe that appearance optimization is more sensitive to the choice of the negative prompt compared to geometry optimization.
For scene stylization, we use a consistent style prompt for editing the appearance of not only all objects but also generating the background panorama, which is achieved through depth-guided ControlNet~\cite{zhang2023adding}.

\subsection{VFX Editing}
The object meshes reconstructed by our method feature detailed UV maps, making them compatible with common 3D software and enabling diverse and photorealistic VFX editing. We implement our VFX editing in Blender, as demonstrated in our main paper. More specifically,
\begin{itemize}
    \item \textit{``\textbf{Freeze it}''} utilizes the Geometry Nodes Modifier and applies a glass material over the original object.
    \item \textit{``\textbf{Ignite it}''} employs the Quick Smoke Effect, setting the \textit{Flow Type} to \textit{Fire} and \textit{Smoke}, with fire color adjustments via Shading Nodes.
    \item \textit{``\textbf{Break it by a ball}''} uses the Cell Fracture Effect to divide the object into multiple fragments, assigning both the object and ball as Rigid Bodies for physics-based simulation in Blender.
\end{itemize}

\section{Comparison with Image-to-3D Method}

\begin{figure}[htbp]
    \centering
    \includegraphics[width=0.95\linewidth]{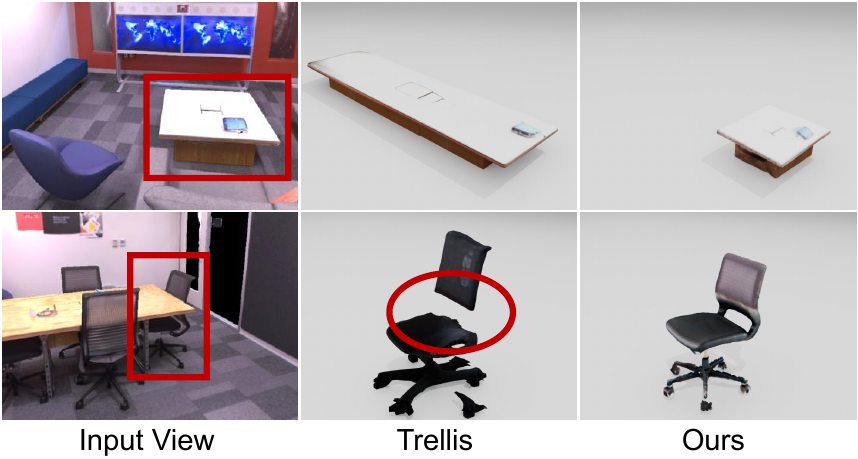}
    \caption{\textbf{Comparison with image-to-3D method Trellis~\cite{xiang2024structured}.} For better visualization, we adjust the location and rotation of Trellis results manually.}
    \label{fig:trellis_supp}
\end{figure}

Unlike decompositional scene reconstruction methods, which recover object geometry along with location, rotation, and scale simultaneously from multi-view images, an alternative approach is to use image-to-3D models for extracting individual objects within a scene.
However, as shown in \cref{fig:trellis_supp}, these models (\eg, Trellis~\cite{xiang2024structured}) face significant challenges in recovering the location, rotation, and scale of objects, with severe performance deterioration under occlusion, making them less applicable for scene reconstruction regardless of views compared to our method.

\begin{figure}[h]
    \centering
    \includegraphics[width=\linewidth]{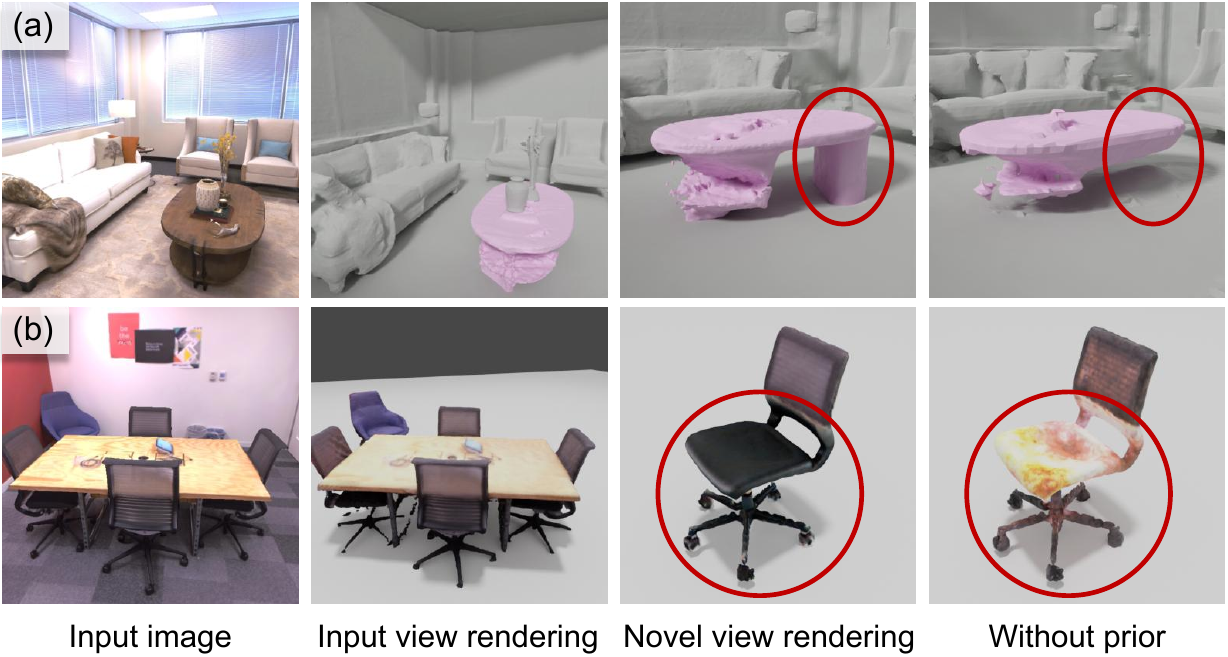}
    \caption{\textbf{Qualitative Examples for Failure Cases.} We present failure cases for both geometry optimization and appearance optimization. The first column displays the input view, while the second and third columns show our results rendered from the input view and a novel view, respectively. The final column provides the corresponding results without applying the geometry prior (top) or appearance prior (bottom), highlighting the improvements introduced by our generative prior.}
    \label{fig:failure_cases}
\end{figure}

\section{Failure Cases and Limitation}
In this section, we present and analyze examples of representative failure.
\cref{fig:failure_cases} (a) demonstrates that our method may produce non-harmonious structures with inaccurate text prompts. For instance, in this example, we use the prompt \textit{``A tea table''}, but the table in this case does not conform to the conventional concept of a tea table. 
A similar issue arises during appearance optimization, as shown in \cref{fig:failure_cases} (b), where we use the prompt \textit{``A black ergonomic chair''}, but the chair in this case is not entirely black—it appears somewhat gray—resulting in a non-harmonious appearance in the completed regions. We believe that leveraging implicit prompts~\cite{gal2022textual} could help alleviate such issues related to text prompts.

Moreover, our method optimizes each object independently in each iteration, using the 3D location information from the reconstruction module alone. This could be further improved by forming functional object groups by composing neighboring objects. Within these groups, \ac{sds} can be employed to optimize inter-object relationships and plausible layouts~\cite{zhou2024gala3d,chen2024comboverse}.
Additionally, \acs{sds}-based methods~\cite{chen2023fantasia3d,qiu2024richdreamer} struggle to reconstruct loose geometries such as grass, hair, sky, and fur, which are challenging to describe with text prompts. In contrast, concurrent methods~\cite{wu2023reconfusion,gao2024cat3d,liu2024reconx,liu20243dgsenhancer} that directly generate novel view images from sparse input views without relying on text prompts may mitigate this limitation. However, these methods often fail to maintain the 3D consistency of objects across views, achieve object-level editing, or reconstruct regions obscured by occlusions.

\end{document}